%% file: acm_mm_video_outpainting.tex
\documentclass[sigconf]{acmart}

\usepackage{subfiles}
\usepackage{subcaption}
\usepackage{sidecap}

\usepackage{multicol}
\usepackage{multirow}
\usepackage{caption}
\usepackage{newfloat}
\usepackage{graphicx}

\usepackage{adjustbox}
\usepackage{dashbox}
\usepackage{xcolor}

\AtBeginDocument{%
  }
    
\copyrightyear{2023}
\acmYear{2023}
\setcopyright{rightsretained}
\acmConference[MM '23]{Proceedings of the 31st ACM International Conference on Multimedia}{October 29-November 3, 2023}{Ottawa, ON, Canada}
\acmBooktitle{Proceedings of the 31st ACM International Conference on Multimedia (MM '23), October 29-November 3, 2023, Ottawa, ON, Canada}
\acmDOI{10.1145/3581783.3612478}
\acmISBN{979-8-4007-0108-5/23/10}

\makeatletter
\gdef\@copyrightpermission{
  \begin{minipage}{0.3\columnwidth}
   \href{https://creativecommons.org/licenses/by/4.0/}{\includegraphics[width=0.90\textwidth]{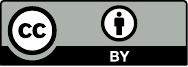}}
  \end{minipage}\hfill
  \begin{minipage}{0.7\columnwidth}
   \href{https://creativecommons.org/licenses/by/4.0/}{This work is licensed under a Creative Commons Attribution International 4.0 License.}
  \end{minipage}
  \vspace{5pt}
}
\makeatother

\begin{document}
\sloppy                         

\title{Hierarchical Masked 3D Diffusion Model for Video Outpainting}

\author{Fanda Fan}
\authornote{Both authors contributed equally to this research while interning at Alibaba Group.}
\email{fanfanda@ict.ac.cn}
\affiliation{%
  \institution{Institute of Computing Technology, Chinese Academy of Sciences}
 \city{Beijing}
 \country{China}
}
\affiliation{%
  \institution{University of Chinese Academy of Sciences}
  \city{Beijing}
  \country{China}
}
\orcid{0000-0002-5214-0959}

\author{Chaoxu Guo}
\authornotemark[1]
\email{chaoxu.guo@nlpr.ia.ac.cn}
\affiliation{%
  \institution{Alibaba Group}
  \city{Beijing}
  \country{China}
}

\author{Litong Gong}
\email{gonglitong.glt@alibaba-inc.com}
\affiliation{%
  \institution{Alibaba Group}
  \city{Beijing}
  \country{China}
}

\author{Biao Wang}
\email{eric.wb@alibaba-inc.com}
\affiliation{%
  \institution{Alibaba Group}
  \city{Beijing}
  \country{China}
}

\author{Tiezheng Ge}
\email{tiezheng.gtz@alibaba-inc.com}
\affiliation{%
  \institution{Alibaba Group}
  \city{Beijing}
  \country{China}
}

\author{Yuning Jiang}
\email{mengzhu.jyn@alibaba-inc.com}
\affiliation{%
  \institution{Alibaba Group}
  \city{Beijing}
  \country{China}
}

\author{Chunjie Luo}
\authornote{Corresponding author.}
\email{luochunjie@ict.ac.cn}
\affiliation{%
  \institution{Institute of Computing Technology, Chinese Academy of Sciences}
  \city{Beijing}
  \country{China}
}

\author{Jianfeng Zhan}
\email{zhanjianfeng@ict.ac.cn}
\affiliation{%
  \institution{Institute of Computing Technology, Chinese Academy of Sciences}
\city{Beijing}
\country{China}
}
\affiliation{%
  \institution{University of Chinese Academy of Sciences}
  \city{Beijing}
  \country{China}
}

\renewcommand{\shortauthors}{Fanda Fan et al.}

\begin{abstract}
Video outpainting aims to adequately complete missing areas at the edges of video frames. Compared to image outpainting, it presents an additional challenge as the model should maintain the temporal consistency of the filled area. In this paper, we introduce a masked 3D diffusion model for video outpainting. We use the technique of mask modeling to train the 3D diffusion model. This allows us to use multiple guide frames to connect the results of multiple video clip inferences, thus ensuring temporal consistency and reducing jitter between adjacent frames. Meanwhile, we extract the global frames of the video as prompts and guide the model to obtain information other than the current video clip using cross-attention. We also introduce a hybrid coarse-to-fine inference pipeline to alleviate the artifact accumulation problem. The existing coarse-to-fine pipeline only uses the infilling strategy, which brings degradation because the time interval of the sparse frames is too large. Our pipeline benefits from bidirectional learning of the mask modeling and thus can employ a hybrid strategy of infilling and interpolation when generating sparse frames. Experiments show that our method achieves state-of-the-art results in video outpainting tasks. More results and codes are provided at our \textcolor{red}{\href{https://fanfanda.github.io/M3DDM/}{project page}}.
\end{abstract}

\ccsdesc[500]{Computing methodologies~Computer vision problems}

\keywords{video outpainting, diffusion model, mask modeling, coarse-to-fine}



\maketitle

\section{Introduction}

\begin{figure*}
    \begin{minipage}[b]{1\linewidth}
        \centering
        \rotatebox[origin=c]{90}{$ratio = 0.6$} \raisebox{-0.4\height}{
        \begin{adjustbox}{precode=\dbox}
            \includegraphics[width=0.55in]{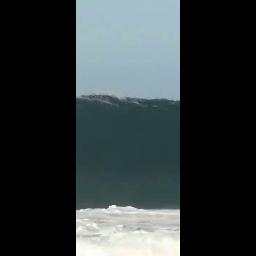}
            \includegraphics[width=0.55in]{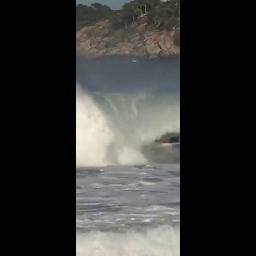}
        \end{adjustbox}
        \begin{adjustbox}{precode=\dbox}
            \includegraphics[width=0.55in]{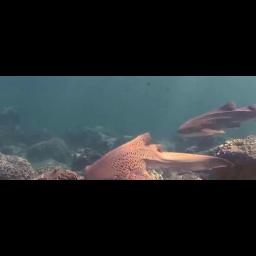}
            \includegraphics[width=0.55in]{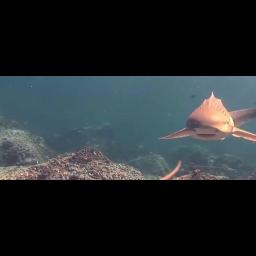}
        \end{adjustbox}
        \begin{adjustbox}{precode=\dbox}
            \includegraphics[width=0.55in]{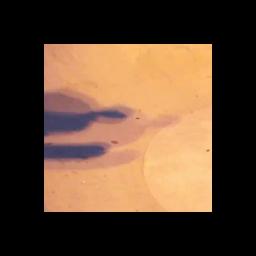}
            \includegraphics[width=0.55in]{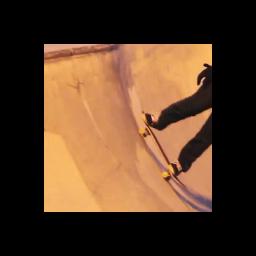}
        \end{adjustbox}}
        \rotatebox[origin=c]{90}{$ratio = 0.5$} \raisebox{-0.4\height}{
        \begin{adjustbox}{precode=\dbox}
            \includegraphics[width=0.55in]{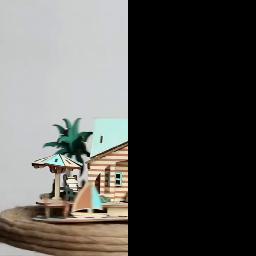}
            \includegraphics[width=0.55in]{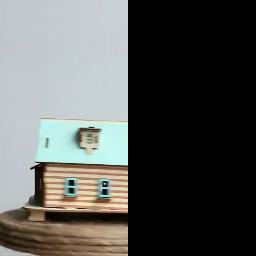}
        \end{adjustbox}
        \begin{adjustbox}{precode=\dbox}
            \includegraphics[width=0.55in]{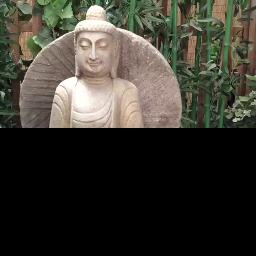}
            \includegraphics[width=0.55in]{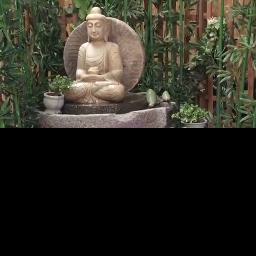}
        \end{adjustbox}}
    \end{minipage}
    \\
    \begin{minipage}[b]{1\linewidth}
        \begin{minipage}[b]{0.01\linewidth}
            \rotatebox[origin=c]{90}{Short Videos} 
        \end{minipage} \raisebox{-0.45\height}{
        \begin{minipage}[b]{0.99\linewidth}
            \centering
            \includegraphics[width=0.6in]{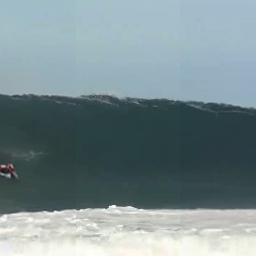}
            \includegraphics[width=0.6in]{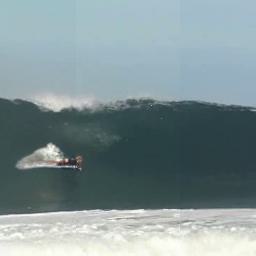}
            \includegraphics[width=0.6in]{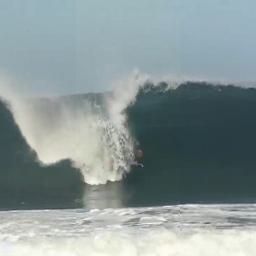}
            \includegraphics[width=0.6in]{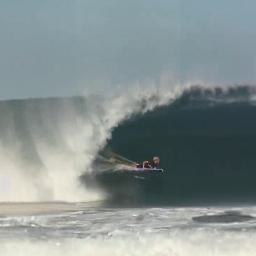}
            \includegraphics[width=0.6in]{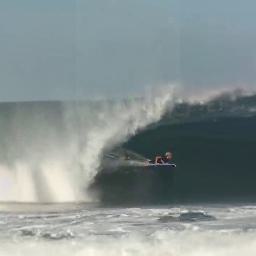}
            \includegraphics[width=0.6in]{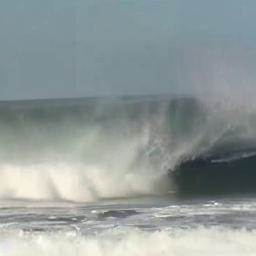}
            \includegraphics[width=0.6in]{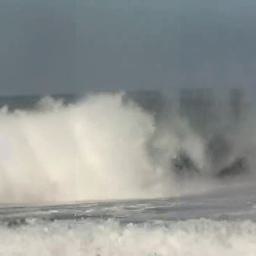}
            \includegraphics[width=0.6in]{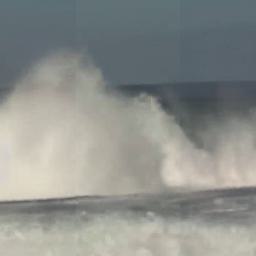}
            \includegraphics[width=0.6in]{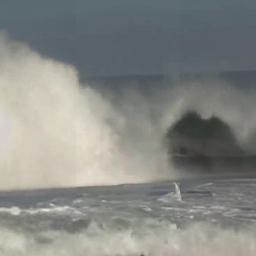}
            \includegraphics[width=0.6in]{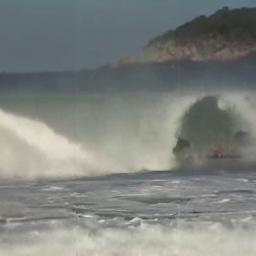}
            \includegraphics[width=0.6in]{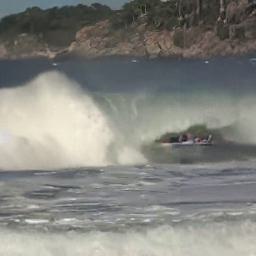}
            \\
            \includegraphics[width=0.6in]{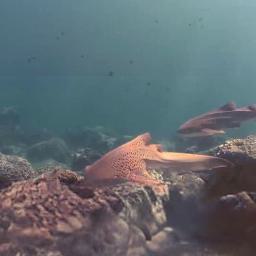}
            \includegraphics[width=0.6in]{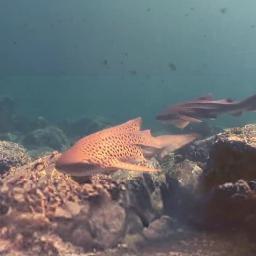}
            \includegraphics[width=0.6in]{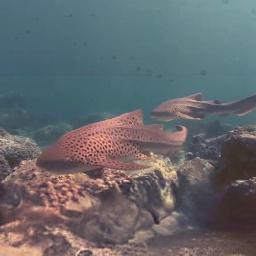}
            \includegraphics[width=0.6in]{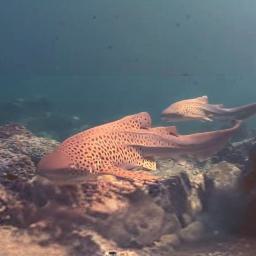}
            \includegraphics[width=0.6in]{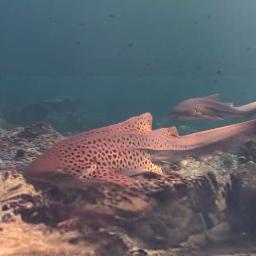}
            \includegraphics[width=0.6in]{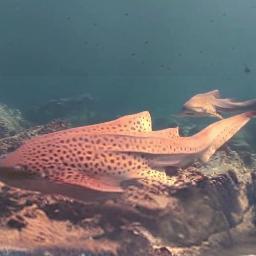}
            \includegraphics[width=0.6in]{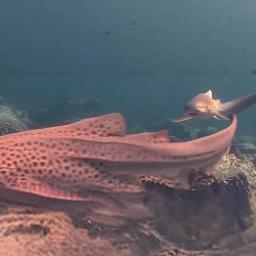}
            \includegraphics[width=0.6in]{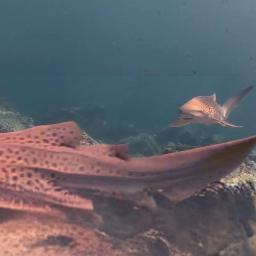}
            \includegraphics[width=0.6in]{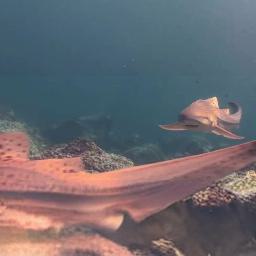}
            \includegraphics[width=0.6in]{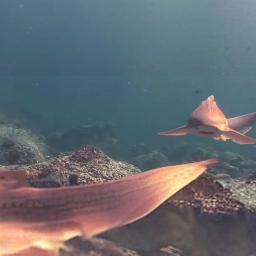}
            \includegraphics[width=0.6in]{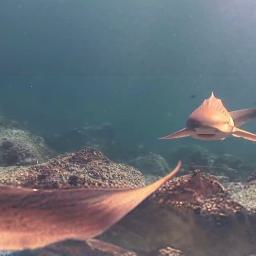}
            \\
            \includegraphics[width=0.6in]{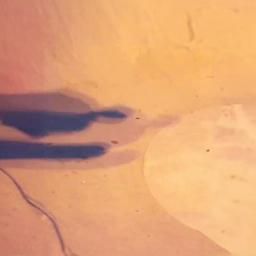}
            \includegraphics[width=0.6in]{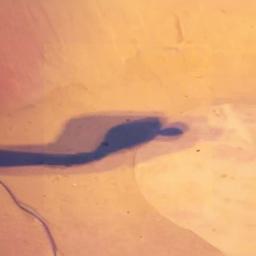}
            \includegraphics[width=0.6in]{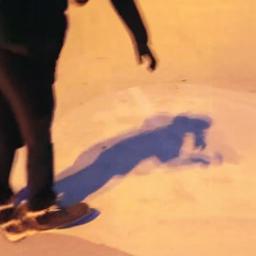}
            \includegraphics[width=0.6in]{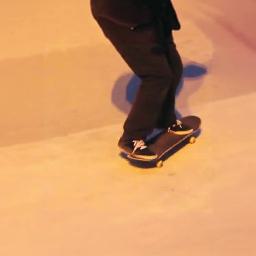}
            \includegraphics[width=0.6in]{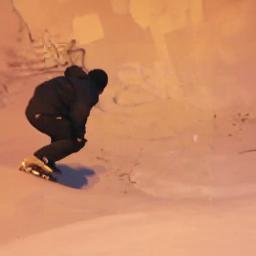}
            \includegraphics[width=0.6in]{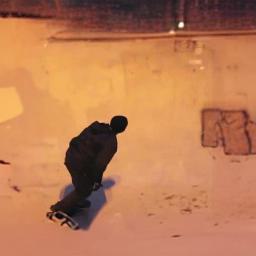}
            \includegraphics[width=0.6in]{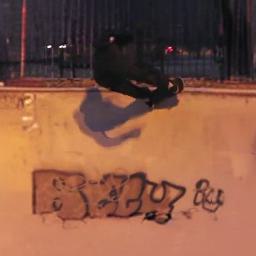}
            \includegraphics[width=0.6in]{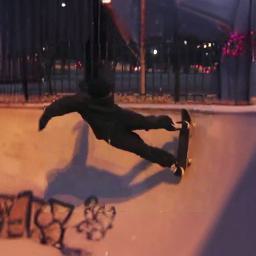}
            \includegraphics[width=0.6in]{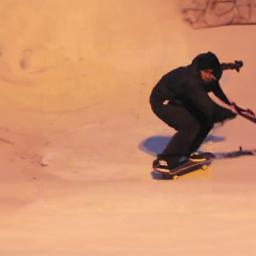}
            \includegraphics[width=0.6in]{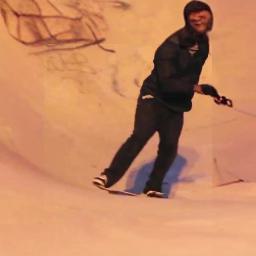}
            \includegraphics[width=0.6in]{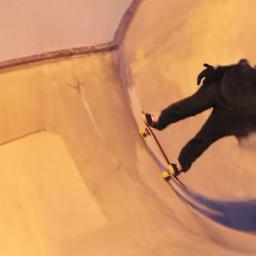}
        \end{minipage}}
    \end{minipage}
    \\
    \noindent\rule{\textwidth}{0.3pt}
    \begin{minipage}[b]{1\linewidth}
        \begin{minipage}[b]{0.01\linewidth}
            \rotatebox[origin=c]{90}{Long Videos} 
        \end{minipage} \raisebox{-0.5\height}{
        \begin{minipage}[b]{0.99\linewidth}
            \centering
            \includegraphics[width=0.6in]{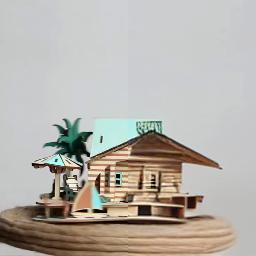}
            \includegraphics[width=0.6in]{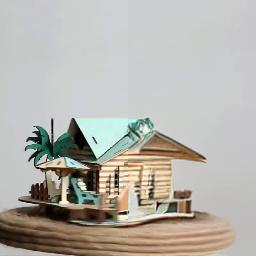}
            \includegraphics[width=0.6in]{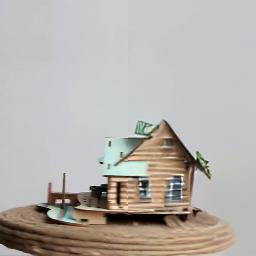}
            \includegraphics[width=0.6in]{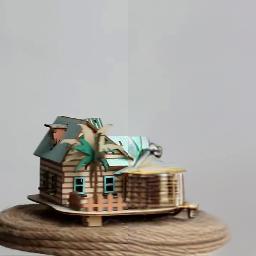}
            \includegraphics[width=0.6in]{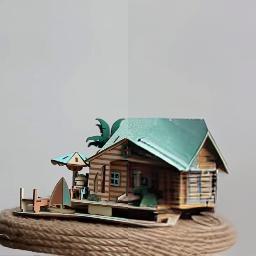}
            \includegraphics[width=0.6in]{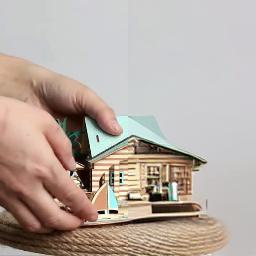}
            \includegraphics[width=0.6in]{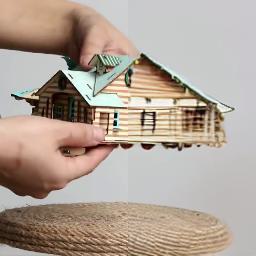}
            \includegraphics[width=0.6in]{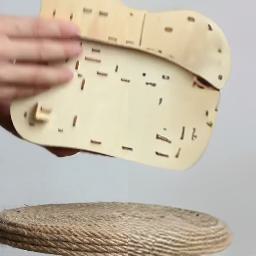}
            \includegraphics[width=0.6in]{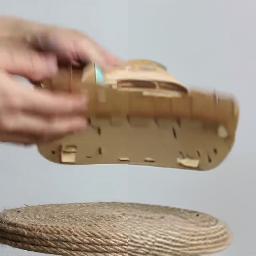}
            \includegraphics[width=0.6in]{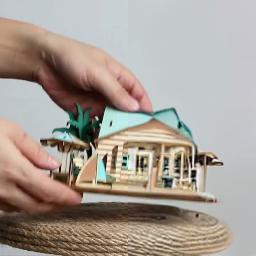}
            \includegraphics[width=0.6in]{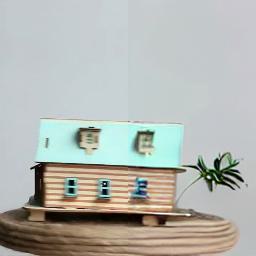}
            \\
            \includegraphics[width=0.6in]{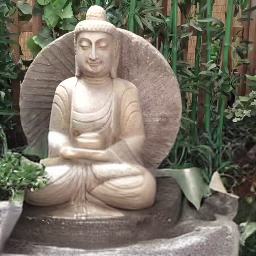}
            \includegraphics[width=0.6in]{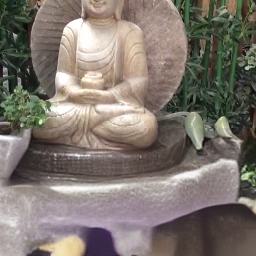}
            \includegraphics[width=0.6in]{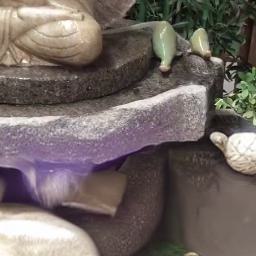}
            \includegraphics[width=0.6in]{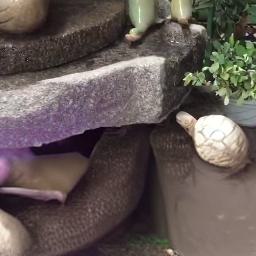}
            \includegraphics[width=0.6in]{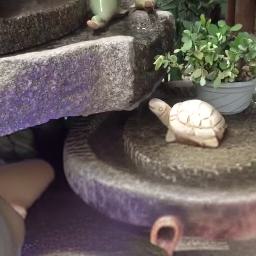}
            \includegraphics[width=0.6in]{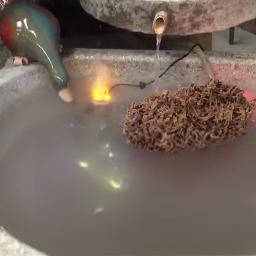}
            \includegraphics[width=0.6in]{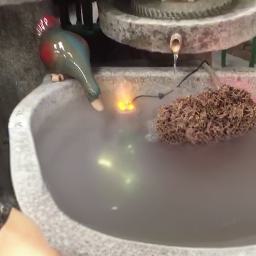}
            \includegraphics[width=0.6in]{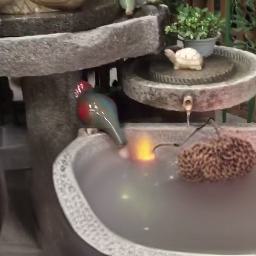}
            \includegraphics[width=0.6in]{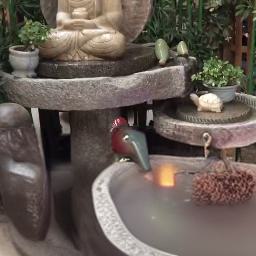}
            \includegraphics[width=0.6in]{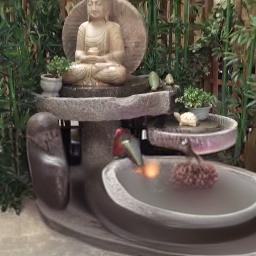}
            \includegraphics[width=0.6in]{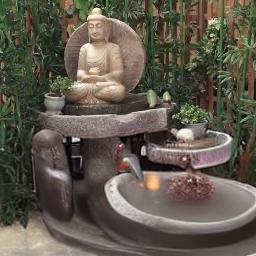}
        \end{minipage}}
    \end{minipage}
    \caption{We propose a Masked 3D Diffusion Model (M3DDM) and a coarse-to-fine inference pipeline for video outpainting.
    Our method can not only generate high temporal consistency and reasonable outpainting results but also alleviate the problem of artifact accumulation in long video outpainting.
    The top row shows the first and last frames of five video clips.
    Each row below shows the video outpainting results of our method.}\label{fig:introfig}
\end{figure*}

The task of video outpainting is to expand edge areas of videos according to the provided contextual information~(the middle part of the videos). In recent years, image outpainting~\cite{cheng2022inout,lin2021edge,wang2021sketch,yang2022scene,chang2022maskgit,saharia2022palette,ramesh2022hierarchical} has been heavily researched and has yielded very promising results with the advent of GAN(Generative Adversarial Network) and Diffusion Model. However, video outpainting is currently far from achieving ideal results. Different from image outpainting, which only considers the spatial appearance of a single image, video outpainting requires the modeling of motion information to ensure temporal consistency among video frames. Besides, videos in real scenarios are typically longer than 5 seconds. It poses two extra challenges: 1) a video would be divided into multiple clips due to the long duration and memory constraints of GPUs. It is challenging to ensure the temporal consistency of generated content among different clips of the same video. and 2) long video outpainting suffers from artifact accumulation issues and meanwhile requires a large amount of computation resources.

A few studies have investigated video outpainting. Dehan~\cite{dehan2022complete} formed a background estimation using video object segmentation and video inpainting methods, and temporal consistency is ensured by introducing optical flow~\cite{dosovitskiy2015flownet,teed2020raft}. However, they often produce poor results in scenarios with complex camera motion and when foreground objects leave the frame. MAGVIT~\cite{yu2022magvit} proposed a generic mask-based video generation model that can also be used for video outpainting tasks. They introduced a 3D-Vector-Quantized~(3DVQ) tokenizer to quantize a video and design a transformer for multi-task conditional masked token modeling. Such a method is able to generate a reasonable short video clip, but the complete result, consisting of multiple clips for a long video, would become poor.  The reason is that it lacks the ability to achieve high temporal consistency in the complete video and suffers from artifact accumulation in multiple clip inferences.

In this work, we focus on video outpainting tasks. To address the issues above, we propose a masked 3D diffusion model~(M3DDM) and a hybrid coarse-to-fine inference pipeline. Recently, the diffusion model~\cite{ho2020denoising,nichol2021improved,dhariwal2021diffusion} has achieved impressive results in image synthesis~\cite{ramesh2022hierarchical,gu2022vector,saharia2022palette} and video generation~\cite{singer2022make,ho2022imagen,blattmann2023videoldm}. Our video outpainting method is based on the latent diffusion models (LDMs)~\cite{rombach2022high}. There are two benefits to choosing LDMs here: 1) They encode the video frames in the latent space instead of the pixel space, thus requiring less memory and achieving better efficiency. 2) Pre-trained LDMs provides good prior about the natural image content and structure that can help our model quickly converges in video outpainting task.

To ensure high temporal consistency in a single clip and across different clips of the same video, we employ two techniques: 1) Masked guide frames, which help to generate current clips that are more semantically coherent and have less jitter with neighboring clips. Mask modeling has proven to be effective in image~\cite{chang2022maskgit} and video generation~\cite{chang2022maskgit,gupta2022maskvit}. During the training phase, we randomly replace the contextual information with raw frames, which have edge areas and act as guide frames. In this way, the model can predict the edge areas not only based on contextual information but also based on adjacent guide frames. The adjacent guide frames can help to generate more coherent and less jittery results. During the inference phase, we iteratively and sparsely outpaint the frames, which allows us to use previously generated frames as guide frames. There are two benefits to using the mask modeling approach. On the one hand, the bidirectional learning mode of mask modeling allows the model to perceive contextual information better, resulting in better single-clip inference. On the other hand, it enables us to use a hybrid coarse-to-fine inference pipeline. The hybrid pipeline not only uses the infilling strategy with the first and last frames as the guide frames but also uses the interpolation strategy with multiple intermediate frames as the guide frames. 2) Global video clips as prompts, which uniformly extracts $g$ global frames from the complete video, encodes them into a feature map using a lightweight encoder, and then interacts with the context of the current video clip (the middle part of the video clip) through cross-attention. This technique enables the model to obtain some global video information when generating the current clip. It is worth noting that the global frames of the video we input do not include the edge areas to be filled in order to \textbf{avoid leakage}. Our experiments show that in scenes with complex camera motion and foreground objects moving back and forth, our method can generate a more temporally consistent complete video.
Some results generated by our method can be seen in Fig.~\ref{fig:introfig}.

\begin{figure}
    \includegraphics[width=\linewidth]{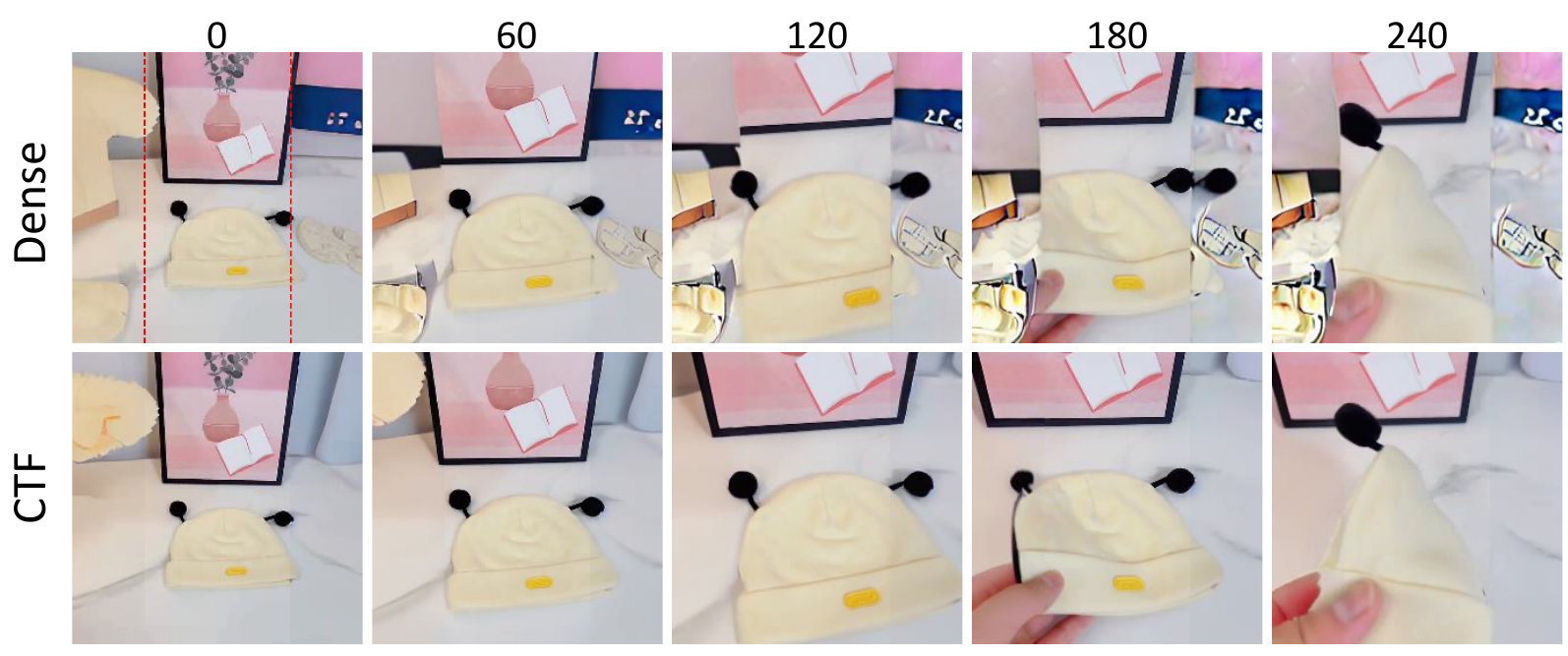}
    \caption{Artifact accumulation problem in long video outpainting.
    We compare two inference methods by our M3DDM: dense and coarse-to-fine~(CTF) inferences.
    The index of the video frame is labeled above the image.
    This case shows horizontal video outpainting with a mask ratio of 0.5. We mark the area to be extended with a red line in the first image.}\label{fig:longvideoproblem}
\end{figure}

Our hybrid coarse-to-fine inference pipeline can alleviate the artifact accumulation problem in long video outpainting. Due to the iterative generation using the guide frames at the inference phase, a bad case generated in the previous step would pollute the subsequent generation results~(This is shown in Fig.~\ref{fig:longvideoproblem}. We will detail later). For the task of long video generation, the coarse-to-fine inference pipeline~\cite{he2022latent,yin2023nuwa} has been proposed recently. In the coarse phase, the pipeline first sparsely generates the keyframes of the video. After that, it generates each frame densely according to the keyframes. Compared to generating the video in a dense manner directly, the coarse stage requires fewer iterations~(because of sparse), thereby alleviating the problem of artifact accumulation in long videos. The existing coarse-to-fine inference pipeline ~\cite{he2022latent,yin2023nuwa} used a three-level hierarchical structure. However, it used only the infilling strategy with the first and last frames to guide the video generation from coarse to fine. This strategy results in a large time interval between key frames generated in the coarsest stage~(the first level), thus bringing degradation in the generated results~(This is shown in Fig.~\ref{fig:effect_interval}.). We also use the coarse-to-fine inference pipeline for video outpainting. Thanks to the masking strategy during the training phase, we can hybridize the infilling strategy and the interpolation strategy together. That means we can not only use the first and last frames as guides for the three-level coarse-to-fine structure but also use multiple frames interpolation to generate the video. Experiments show that our hybrid coarse-to-fine inference pipeline brings lower artifacts and better results in long video generation.

Our main contributions are as follows:
\begin{itemize}
\item To the best of our knowledge, we are the first to use a masked 3D diffusion model for video outpainting and achieve state-of-the-art results.
\item We propose a bidirectional learning method with mask modeling to train our 3D diffusion model.
Additionally, we show that using guide frames to connect different clips of the same video can effectively generate video outpainting results with high temporal consistency and low jitter.
\item We extract global temporal and spatial information as prompt from global frames of the video and feed it into the network in the form of cross-attention, which guides the model to generate more reasonable results.
\item We propose a hybrid coarse-to-fine generation pipeline that combines infilling and interpolation when generating sparse frames. Experiments show that our pipeline can reduce artifact accumulation in long video outpainting while maintaining a good level of temporal consistency.
\end{itemize}

\section{Related Work}
This section introduces the related diffusion model, mask modeling, and the Coarse-to-Fine pipeline.

\textbf{Diffusion Model.}
The diffusion model~\cite{sohl2015deep,ho2020denoising,nichol2021improved} has recently become the best technology in image generation~\cite{ramesh2022hierarchical,saharia2022palette}, especially in video generation~\cite{singer2022make,ho2022imagen,molad2023dreamix}.
Compared with GAN~\cite{goodfellow2020generative}, it can generate samples with richer diversity and higher quality~\cite{dhariwal2021diffusion}.
Considering the significant achievements of the diffusion model in video generation, we adopt it as the main body of our video outpainting method.
LDMs~\cite{rombach2022high} are diffusion models in the latent space, which reduce the GPU memory usage, and their open-source parameters are excellent image priors for our video outpainting task.

\textbf{Mask Modeling.}
Mask modeling was first proposed in the BERT~\cite{devlin2018bert} in the field of NLP for language representation learning.
BERT randomly masks tokens in sentences and performs bidirectional learning by predicting the masked tokens based on context.
MAE~\cite{he2022masked} has demonstrated that mask modeling can be effectively used in unsupervised image representation learning in the field of computer vision.
This is achieved by masking patch tokens in the image and predicting the original patch tokens based on context.
Recently, Mask modeling has also been used in the field of video generation~\cite{gupta2022maskvit}.
In more recent times, the combination of mask modeling and diffusion model has been applied to image~\cite{gu2022vector,wei2023diffusion} and video generation~\cite{voleti2022masked} tasks.
In this paper, we do not apply masks on images or entire frames of videos, but rather, in consideration of the feature of video outpainting, masks are applied to the surrounding areas of the video that need to be filled with a probability.
Our experiments show that for video outpainting tasks, the employment of the diffusion model technique with mask modeling can generate higher-quality results.

\begin{figure*}[h]
    \centering
    \includegraphics[width=1\linewidth]{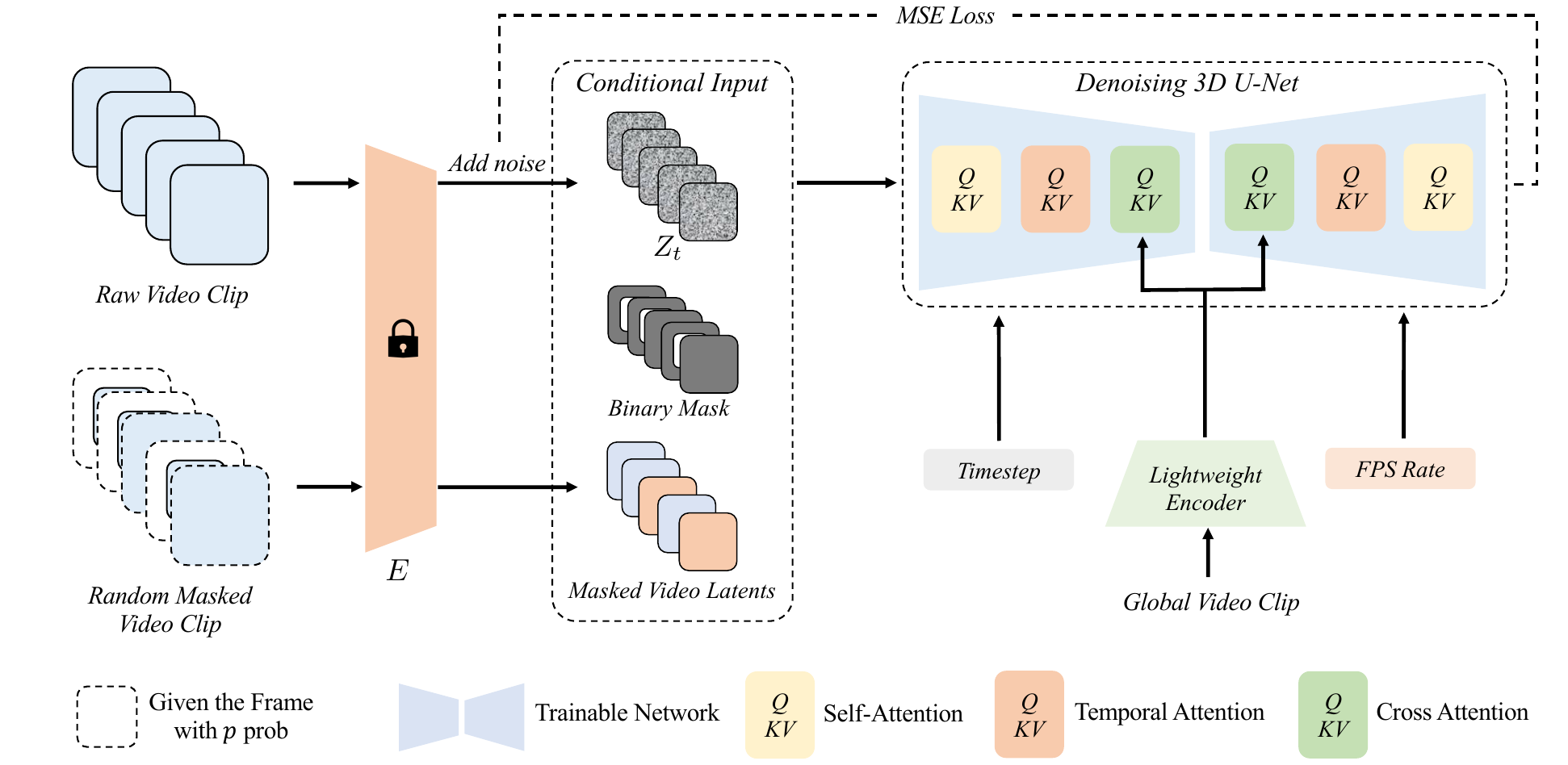}
    \caption[width=\linewidth]{Masked 3D Diffusion Model Framework. During training, we concatenate corrupted raw video latents, random masked video latent, and masks before feeding them into the 3D UNet network. The network predicts the noise in the corrupted raw latents, allowing us to calculate the MSE loss with the added noise. Additionally, we uniformly select $g$ global frames from the video as a prompt and feed them into a trainable video encoder. Then the global frames feature map is placed in the cross-attention module of the 3D UNet.
    }\label{fig:framework}
\end{figure*}

\textbf{Coarse-to-Fine Pipeline.}
In the generation of long videos, models often suffer from artifact accumulation due to the auto-regressive strategy.
For the method of generating videos with guidance frames, artifacts from the previous video clips often affect the later iterations.
Recent research~\cite{he2022latent, yin2023nuwa, blattmann2023videoldm} adopt a coarse-to-fine generation pipeline for video generation.
They first generate sparse key frames of the video and alleviate the artifact problem by reducing the number of iterations.
In our video outpainting task, we adopt the coarse-to-fine inference pipeline and use both infilling strategies with two guidance frames and interpolation strategies with multiple guidance frames to help alleviate the problem of artifact accumulation in long videos.



\section{Methodology}

\subsection{Preliminaries}

Diffusion models~\cite{sohl2015deep,ho2020denoising,nichol2021improved,dhariwal2021diffusion} are probabilistic models that learn the data distribution $p_{data}$ by first forward adding noise to the original distribution, and then gradually denoising the normal distribution variables to recover the original distribution.
In the forward noising process, a sample $x_0$ can corrupted from $t = 0$ to
$t = T$ using the following transition kernel:
\begin{equation}
q_t(x_t|x_{t-1}) = \mathcal N(x_t; \sqrt{1 - \beta_t} x_{t-1}, \beta_t I).
\end{equation}
And $x_t$ can be directly sampled from $x_0$ using the following accumulation kernel:
\begin{equation}
x_t = \sqrt{\widetilde \alpha_t} x_0 + \sqrt{1 - \widetilde \alpha_t} \epsilon, 
\end{equation}
where $\widetilde \alpha_t = \prod_{s = 1}^t (1 - \beta_s)$, and $\epsilon \sim \mathcal N(0, 1)$.
In the process of denoising, a deep model is typically trained to predict the noise in a corrupted signal $x_t$.
The loss function of the model can be simply written as 
\begin{equation}
\label{eq:loss_func_prim}
L_{DM} = \mathbb{E}_{x, \epsilon \sim \mathcal{N}(0, 1), t} [{\lVert \epsilon - \epsilon_\theta (x_t, c, t) \rVert}_2^2], 
\end{equation}
where $c$ is the conditional input and $t$ is uniformly sample from $\{1, \ldots, T\}$.

LDMs~\cite{rombach2022high} additionally trained an encoder $E$ to map the original $x_0$ from the pixel space to the latent space, greatly reducing memory usage and making the model more efficient with an acceptable loss.
Then, the decoder $\mathcal{D}$ is used to map $z_0$ back to the pixel space.
Considering that video outpainting task requires large memory, we choose the LDMs framework as our pipeline.
Additionally, the pre-training parameters of LDMs can serve as a good image prior, which helps our model converge faster.
In equation~\ref{eq:loss_func_prim}, we rewrite $x$ as $z$.

\subsection{Masked 3D Diffusion Model}

With the help of LDMs, a naive approach is to concatenate the noisy latent of raw video clip with the context of the video clip as a conditional input and train a model to predict the added noise. Thus, the model can recover the raw video clip~(the original video) from the randomly sampled Gaussian noise distribution. Since videos usually contain hundreds of frames, the model is required to perform inference on different clips of the same video separately, and then the generated clips are stitched together to form the final outpainting result of the complete video. Under this circumstance, the naive approach above cannot guarantee the temporal consistency of the predicted video clips.\label{sec:sdm}

To address it, we propose the masked 3D diffusion model, whose overview is shown in Fig.~\ref{fig:framework}. Our model can generate F frames at once. We describe our network architecture in Appendix~\ref{sec:netarch}. We sample video frames with different frames per second ~(fps) and additionally feed the fps into 3D UNet. This allows us to use one unifying model to adapt to videos with different frame rates. Our framework follows LDMs and first maps video frames in the pixel space to the latent space through a pre-trained encoder $E$. At the training stage, each context frame is replaced with raw video frames with a probability $p_{frame}$ before they are fed into the encoder $E$. Therefore, our model has the ability to use guide frames at the inference stage, and more than two frames can be conditioned to facilitate the generation of other frames. This modification has two benefits. First, it enables our coarse-to-fine inference pipeline, ensuring consistent inference time across multiple passes. Second, compared to solely using the first or the last raw frames as input conditions, bidirectional learning can help the model better perceive contextual information, thereby improving generation quality. We would validate this point in our ablation study.

\subsubsection{Mask Strategy}
\label{sec:maskcase}
In order to construct the training samples for video outpainting, we randomly mask out the edges of each frame. We mask a frame with different direction strategies: four-direction, single-direction, bi-direction (left-right or top-down), random in any of four directions, and mask all. Taking into account the practical application scenarios, we adopt the proportions of these five strategies as 0.2, 0.1, 0.35, 0.1, and 0.25, respectively. The "mask all" strategy enables the model to perform unconditional generation, which allows us to adopt the classifier-free guidance \cite{ho2022classifier} technique during the inference phase. Considering the size of the edge area that needs to be outpainted in practical application scenarios, we randomly sample the mask ratio of a frame from $[0.15, 0.75]$ uniformly.

In order to generate masked guide frames, we replace the contextual frame with the raw frame in three cases:
1) All F frames are given only context information, where each frame is masked with the above masking strategy.
2) The first frame or the first and last frames of F frames are replaced with the unmasked raw frame, and the rest of the frames are given only context information.
3) Any frame is replaced with an unmasked raw frame with probability $p_{frame} = 0.5$.
The guide frames allow the model to predict the edge areas not only based on contextual information but also based on the adjacent guide frames. The adjacent guide frames can help to generate more coherent and less jittery results. We evenly distribute the training proportions of the three cases. The proportions of these three cases are 0.3, 0.35, and 0.35, respectively. We do not only train using case 3 because we considered that the first two cases would be used more frequently during the prediction phase.

\subsubsection{Global Video Clip as a Prompt}
In order to enable the model to perceive global video information beyond the current clip, we uniformly sample $g$ frames from the video. These global frames are passed through a learnable lightweight encoder to obtain the feature map, which is then fed into 3D-UNet via cross-attention. We do not feed the global frames in the input layer of 3D-UNet because we suggest that cross-attention can help masked frames interact with global frames more thoroughly. It is worth noting that the global frames passed in here are aligned with the context of the current video clip and are also masked in the same way as other frames to avoid information leakage.

\subsubsection{Classifier-free Guidance}
Classifier-free guidance~\cite{ho2022classifier} has been proven to be effective in diffusion models. Classifier-free guidance improves the results of conditional generation, where the implicit classifier $p_\theta(c|z_t)$ assigns high probability to the conditioning $c$.
In our case, we have two conditional inputs. One is the context information of the video $c_1$, and the other is the global video clip $c_2$. We jointly train the unconditional and conditional models by randomly setting $c_1$ and $c_2$ to a fixed null value $\varnothing$ with probabilities $p_1$ and $p_2$. At inference time, we follow Brooks'~\cite{brooks2022instructpix2pix} approach for two conditional inputs and use the following linear combination of the conditional and unconditional score estimates:
\begin{equation}
    \begin{split}
        \hat{\epsilon}(z_t, c_1, c_2) = \epsilon(z_t, \varnothing, \varnothing) + s_1(\epsilon(z_t, c_1, \varnothing) - \epsilon(z_t, \varnothing, \varnothing)) \\
        + s_2(\epsilon(z_t, c_1, c_2) - \epsilon(z_t, c_1, \varnothing)), 
    \end{split}
\end{equation}
where $s_1$ and $s_2$ are the guidance scales.
The guidance scales control whether the generated video relies more on the context of the video or on the global frames of the video.

\begin{figure}[t]
    \centering
    \includegraphics[width=\linewidth]{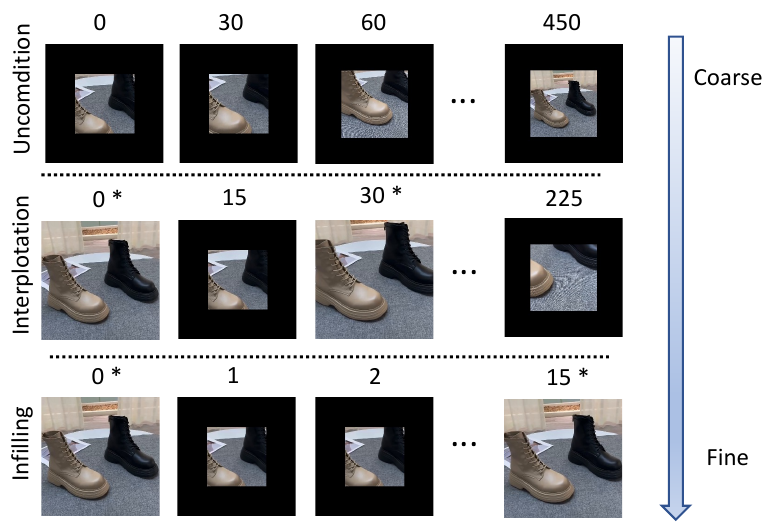}
    \caption[width=\linewidth]{Coarse-to-Fine Pipeline.
    Our model can generate 16 frames at a time.
    We label the index above each frame,
    and those with $*$ indicate that the result has already been generated in the previous step and used as a conditional input for the model in the current step. Our pipeline includes a hybrid strategy of infilling and interpolation.}\label{fig:coarse-to-fine-pipeline}
\end{figure}

\begin{figure*}
    \includegraphics[width=\linewidth]{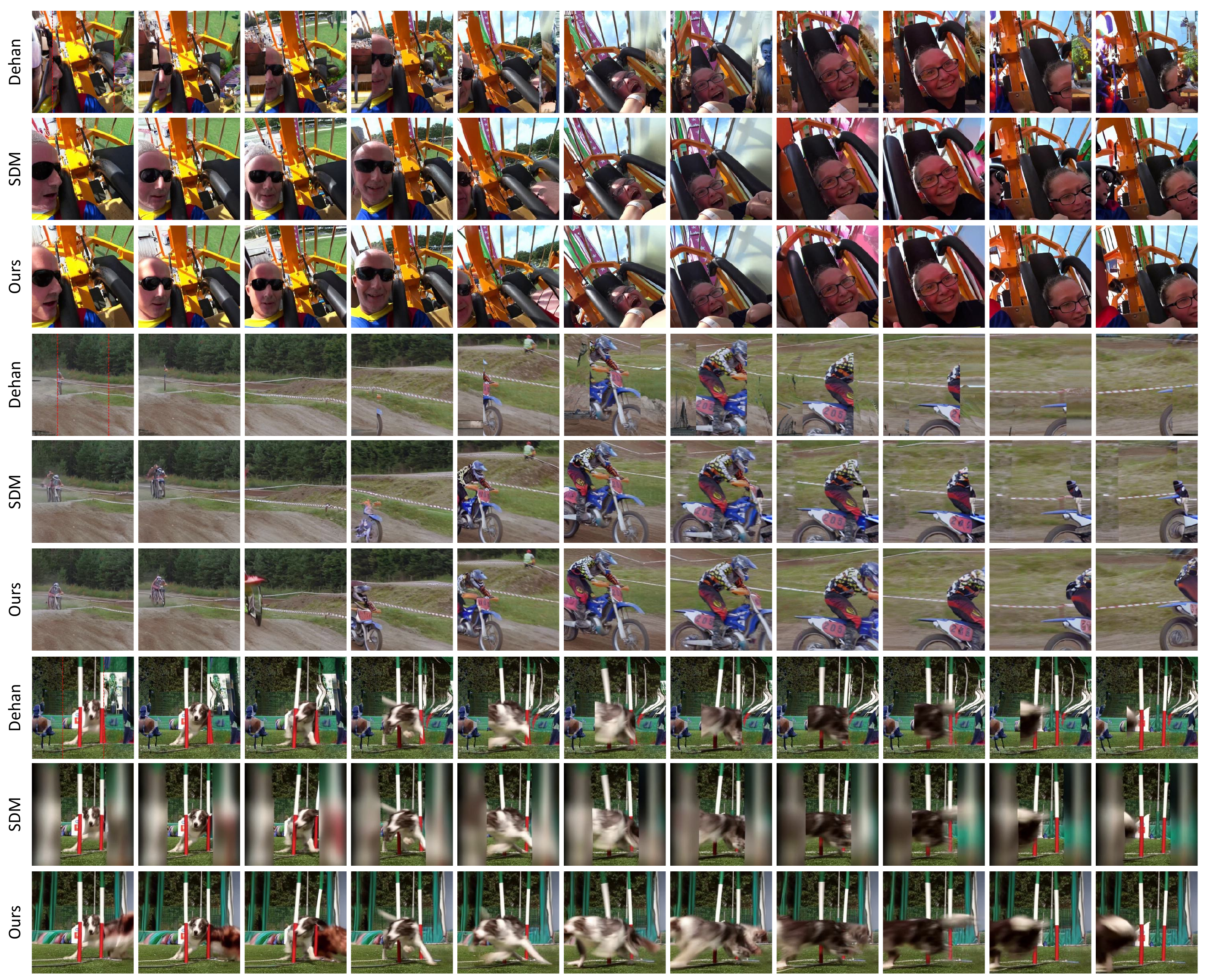}
    \caption{Qualitative Comparison of short video outpainting. We present the results of three groups of horizontally oriented video outpainting with ratio proportions of 0.4, 0.5, and 0.6. We mark the area to be extended with a red line in the first image.}\label{fig:shortvideooutpainting}
\end{figure*}

\begin{figure}
    \begin{minipage}[b]{0.45\linewidth}
        \centering
        \includegraphics[width=1\linewidth]{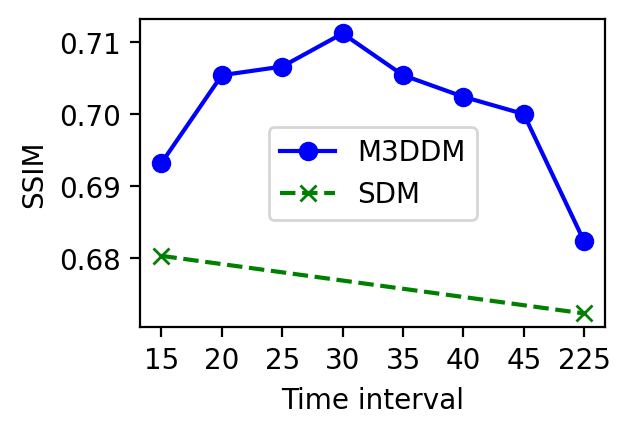}
        \subcaption{}\label{fig:effect_interval}
    \end{minipage}
    \begin{minipage}[b]{0.45\linewidth}
        \centering
        \includegraphics[width=1\linewidth]{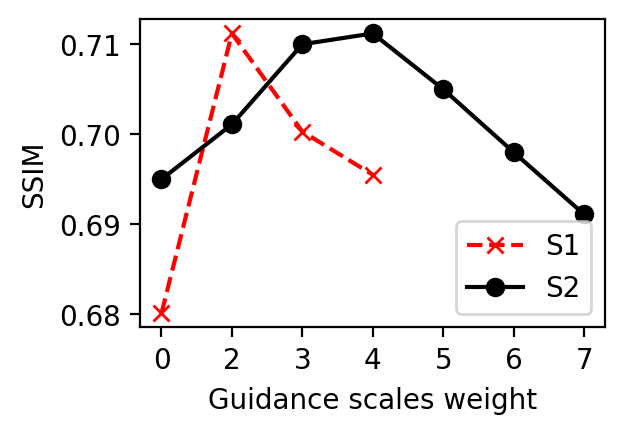}
        \subcaption{}\label{fig:effect_time}
    \end{minipage}
    \caption{Evaluation of different time intervals and guidance scale weights.}
\end{figure}

\begin{figure*}
    \includegraphics[width=\linewidth]{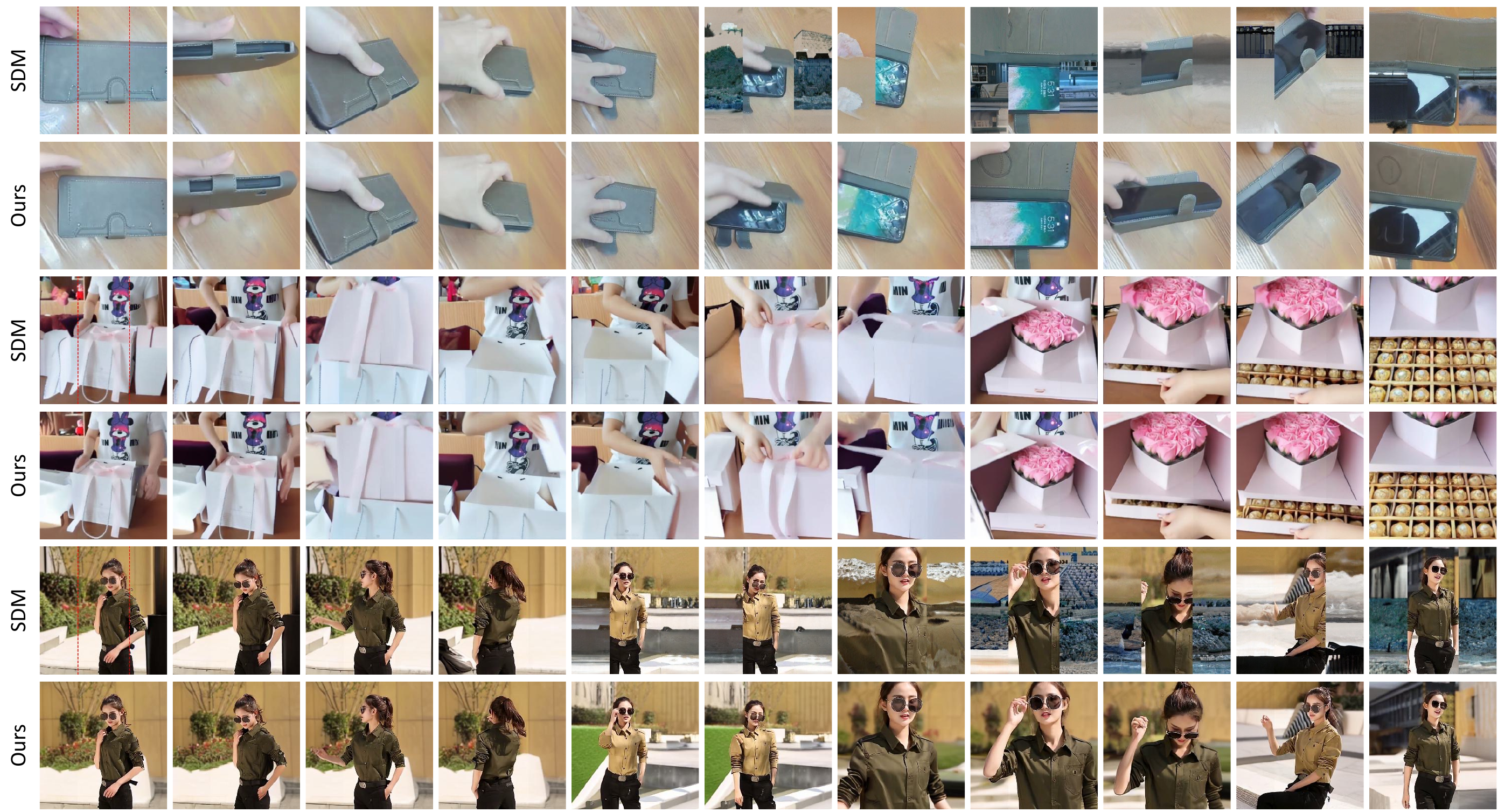}
    \caption{Qualitative Comparison of long video outpainting. We present the results of three groups of horizontally oriented video outpainting with a ratio proportion of 0.6. We mark the area to be extended with a red line in the first image.}\label{fig:longvideooutpainting}
\end{figure*}

\subsection{Hybrid Coarse-to-Fine Pipeline for Video Outpainting}
In video generation tasks, the generation of long videos often leads to the accumulation of artifacts, resulting in degraded performance.
Recent research~\cite{yin2023nuwa,he2022latent,blattmann2023videoldm} used a hierarchical structure first to generate sparse key frames of the video, and then use an infilling strategy to fill in dense video frames.
The infilling strategy requires the first and last frames as guide frames to guide the generation of the next level. However, using infilling alone can result in a large time interval between frames in the coarse phase.
For example, as shown in Fig.~\ref{fig:coarse-to-fine-pipeline}, if we only use infilling strategy, our model requires a frame interval of 225 instead of 30 in the coarsest level.
Due to the difficulty of the problem and the lack of long video data in the training set, this can lead to poor results.

Thanks to bidirectional learning, our 3D UNet can perform video outpainting by combining infilling and interpolation. This avoids the problem of large frame intervals in the coarse generation phase.
Our coarse-to-fine process diagram is shown in Fig.~\ref{fig:coarse-to-fine-pipeline}.
Our coarse-to-fine pipeline is divided into three levels.
In the first level (coarse), we unconditionally generate the first video clip and then iteratively generate all keyframes based on the results of the last frame from the previous iteration.
In the second level (coarse), we use the keyframes generated in the first level as conditional inputs to generate more keyframes through interpolation.
In the third level (fine), we generate the final video outpainting result with a frame interval of 1, using the first and last frames as guide frames for dense generation.

\section{Experiments}
To verify the effectiveness of our masked 3D diffusion model for video outpainting,
we conduct evaluations on three datasets:
DAVIS~\cite{perazzi2016benchmark}, YouTube-VOS~\cite{xu2018large}, and our 5M E-commerce dataset.
DAVIS and YouTube-VOS are commonly used datasets for video inpainting and outpainting.
However, their average video length is short.
Therefore, to validate the outpainting performance for longer videos, we collect long videos from the e-commerce scene, called 5M E-commerce dataset.
Our 5M E-commerce dataset contains over 5 million videos, with an average video length of around 20 seconds.
It consists of videos provided by advertisers to showcase their products, mainly including furniture, household goods, electronics, clothing, food, and other commodities.
We describe our implementation details in Appendix~\ref{sec:implementation}.

\subsection{Baselines and Evaluation Metrics}
We compare with the following methods: 1) Dehan~\cite{dehan2022complete} proposed a framework for video outpainting. They separated the foreground and background and performed flow estimation and background estimation separately before integrating them into a complete result. 2) We also train a simple diffusion model~(SDM) based on stable diffusion~\cite{rombach2022high} as a baseline. It adopts the first frame and last frame as condition frame concatenated with the context video clip at the input layer without using mask modeling and fed into the denoising 3D UNet. Meanwhile, we do not use global features as a prompt, and cross attention is removed.
3) MAGVIT~\cite{gupta2022maskvit} used mask modeling technology to train a transformer~\cite{dosovitskiy2020image} for video generation in the 3D Vector-Quantized~\cite{van2017neural,esser2021taming} space. We included this set of comparisons in Appendix~\ref{sec:magvitcomp}.

We follow~\cite{dehan2022complete} and use five commonly used evaluation metrics: Mean Squared Error(MSE), Peak Signal To Noise Ratio (PSNR), structural similarity index measure (SSIM)~\cite{wang2004image},  Learned Perceptual Image Patch Similarity (LPIPS)~\cite{zhang2018unreasonable}, and Frechet Video Distance (FVD)~\cite{unterthiner2018towards}.
To evaluate MSE, PSNR, SSIM, and FVD, we convert the generated results into video frames with a value range of $[0, 1]$, while LPIPS is evaluated using a value range of $[-1, 1]$.
For the FVD evaluation metric, we use a uniform sampling of 16 frames per video for evaluation.

\input{table/compare_davis_youtube}

\subsection{Short Video Outpainting}
\subsubsection{Qualitative Comparison.}

In Fig.~\ref{fig:shortvideooutpainting}, we present the results of three methods for horizontal video outpainting.
It can be seen that Dehan~\cite{dehan2022complete}, although capable of generating a better background, produces poor foreground results due to its dependence on the result of flow prediction.
The structural information of the subject in the filling area is essentially lost, resulting in unreasonable outcomes.
SDM, with the help of strong diffusion tools and the addition of guide frames, is able to preserve the spatial structure of the filling area within a short interval.
However, due to the lack of global information, it also loses many reasonable predictions in generating the complete video.
In the third group of results with a mask ratio of 0.6 in Fig.~\ref{fig:shortvideooutpainting}, SDM produces a bad case with some noisy outcomes.
We find that the introduction of mask modeling can alleviate the proportion of bad cases generated by the diffusion model.
We will discuss this further in the ablation study.
As can be seen in our method, we not only preserve the spatial information of the foreground subject in the filling area but also generate a reasonable background.
Thanks to the introduction of global video information, our method can perceive that the motorcycle should appear in the filling area in the third group 3 at an early stage.
Moreover, compared with SDM, our additional mask modeling can generate fewer bad cases.

\subsubsection{Quantitative Results.}

We compare the outpainting results in the horizontal direction on datasets DAVIS and YouTube-VOS with Dehan~\cite{dehan2022complete} and SDM, using mask ratios of 0.25 and 0.666.
For each evaluation metric, we report their mean values across all test samples.
Our evaluation results on the DAVIS and YouTube-VOS datasets are shown in Table~\ref{tab:quantitative1}.

\subsection{Long Video Outpainting}
We demonstrate a comparison between densely prediction and coarse-to-fine (CTF) prediction on a long video in Fig.~\ref{fig:longvideoproblem}.
It can be seen that densely prediction not only produces unreasonable results in the early predictions of the video but also suffers from the accumulation of artifacts from previous iterations.
We claim that the CTF prediction method can generate more reasonable results in the early predictions by considering longer video clip information, while also alleviating the problem of artifact accumulation due to the decrease of times of auto-regressive inference.

\subsubsection{Study of Time Interval Between Frames}
We explore the relationship between the frame interval generated in the coarse stage and the results in Fig.~\ref{fig:effect_interval}.
We randomly select 100 long videos from our 5M e-commerce dataset as the test set.
Interval 15 means a two-level prediction structure, while greater than 15 means a three-level structure.
We found that the results generated by the three-level structure were better than those generated by the two-level structure.
However, further increasing the interval between frames in the third level resulted in performance degradation in the M3DDM and SDM models.
Especially when only using the infilling strategy, a frame interval of 225 resulted in greater degradation in both the SDM and M3DDM.
It is worth noting that SDM can only use a time interval of 225 at the third level because it uses the first and last frames as guide frames.

For qualitative comparison, we contrast our approach with SDM on 3 long videos in our 5M e-commerce dataset. The SDM here adopts a two-level CTF with time intervals of $[15, 1]$ respectively. As shown in Fig.~\ref{fig:longvideooutpainting}, our M3DDM not only generates foreground subjects well in the area to be filled but also produces more consistent background results.

\subsection{Ablation Study}

We conduct an ablation study on our 5M e-commerce dataset.
We randomly select 400 videos from 5M e-commerce dataset, with an average length of 20 seconds.
In our simple diffusion model (SDM), we only use the first and last guide frames concatenation with the context of the video clip for training, without incorporating mask modeling and global frames.
In order to independently verify the improvement effect of mask modeling on the diffusion model, we employ a SDM and combined it with the mask modeling~(As we mentioned in Sec.\ref{sec:maskcase}) to train the masked SDM~(MSDM).
Our approach is to introduce a global video clip as a prompt based on the masked SDM.
In long video inference, we use a two-level coarse-to-fine inference structure on the SDM (three levels have a degradation in performance), and a three-level coarse-to-fine inference pipeline is used in the masked SDM and our approach.
As shown in Table~\ref{tab:ablation}, compared with short videos, our approach and SDM have a larger performance gap in long videos.
Compared with SDM, MSDM produced better video outpainting results.

\input{table/ablation}

\subsubsection{Effective of Guidance Scales}
In Fig.~\ref{fig:effect_time}, we present the effectiveness of guidance scales. When we change $s_1$, we fix $s_2$ at 4. When we change $s_2$, we fix $s_1$ at 2. $s_1$ controls the model to generate results that are more relevant to the video context, and $s_2$ helps the model generate more reasonable results in scenes where the camera is moving or the foreground subject is moving. We found that it is more important to have classifier-free guidance for video context. When we do not have classifier-free guidance for video context, the performance degrades significantly. At the same time, having classifier-free guidance for video context and global frames brings better results.

\section{Conclusion}
In this paper, we propose a 3D diffusion model based on mask modeling for video outpainting.
We use bidirectional learning and globally encoding video frames as a prompt for cross-attention with context. The bidirectional learning approach of mask modeling allows us to have more flexible strategies in the inference stage while better perceiving adjacent frame information. The addition of a global video clip as a prompt further improves our method's performance. In most cases of camera movement and foreground object sliding, global frames help the model generate more reasonable results in filling the areas. We also propose a hybrid coarse-to-fine inference pipeline for video outpainting, which combines infilling and interpolation strategies. Experiments show that our method achieves state-of-art results.


\bibliographystyle{ACM-Reference-Format}
\balance
\bibliography{acm_mm_video_outpainting}

\clearpage

\appendix

\section{Appendix Overview}
Our supplementary materials provide additional experimental results and comparison methods to better evaluate our approach.
At the same time, we also supplement the implementation details that were not expanded in the main text due to space limitations.
Our supplementary materials are described in the following sections:

\begin{itemize}
\item Compared with MAGVIT on Something-Something V.2~(SSv2) Dataset. We additionally conduct a comparative experiment with MAGVIT~\cite{yu2022magvit}.
We directly obtain quantitative results from their paper and compare them using the same setting on the SSv2 dataset.
\item Network architecture and implementation details. 
\item Limitations. We briefly presented some bad cases generated by our method.

\end{itemize}

\section{Compared with MAGVIT}
\label{sec:magvitcomp}
In the introduction of our main text, MAGVIT~\cite{yu2022magvit} has been briefly introduced.
They used mask modeling technology to train a transformer~\cite{dosovitskiy2020image} for video generation in the 3D Vector-Quantized~\cite{van2017neural,esser2021taming} space.
They also evaluated MAGVIT's performance in video outpainting tasks in the paper.
However, MAGVIT lacks constraints on different clips of the same video, resulting in poor temporal consistency in the generated results between different clips.
Our M3DDM model, utilizing the diffusion model and introducing global video frames as prompts, along with mask modeling and guided frame techniques, not only performs well in generating long videos but also surpasses MAGVIT~\cite{yu2022magvit} in short video outpainting.

\begin{figure*}
    \begin{minipage}[b]{1\linewidth}
        \begin{minipage}[b]{0.01\linewidth}
            \rotatebox[origin=c]{90}{GT}
        \end{minipage} \raisebox{-0.5\height}{
        \begin{minipage}[b]{0.99\linewidth}
            \centering
            \includegraphics[width=0.52in]{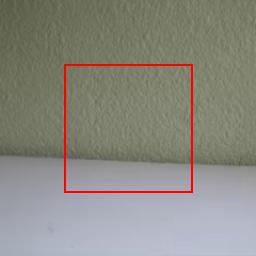}
            \hspace{-1mm}
            \includegraphics[width=0.52in]{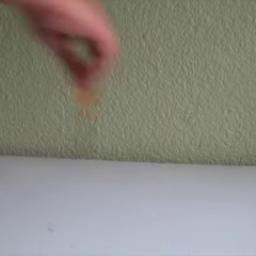}
            \hspace{-1mm}
            \includegraphics[width=0.52in]{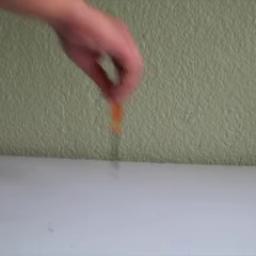}
            \hspace{-1mm}
            \includegraphics[width=0.52in]{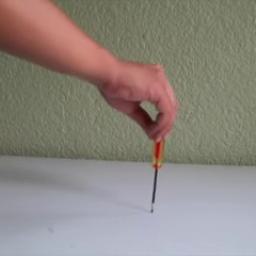}
            \hspace{-1mm}
            \includegraphics[width=0.52in]{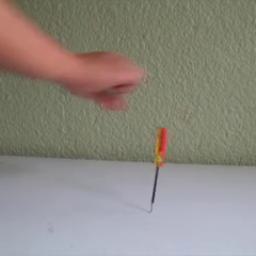}
            \hspace{-1mm}
            \includegraphics[width=0.52in]{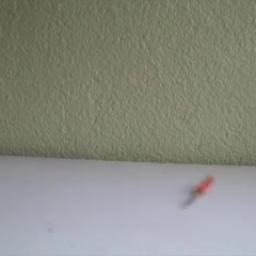}
            \hspace{1mm}
            \includegraphics[width=0.52in]{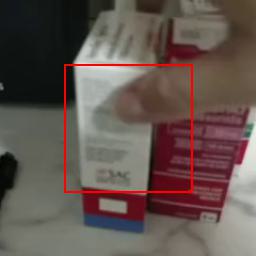}
            \hspace{-1mm}
            \includegraphics[width=0.52in]{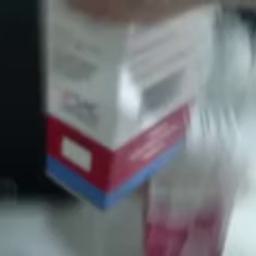}
            \hspace{-1mm}
            \includegraphics[width=0.52in]{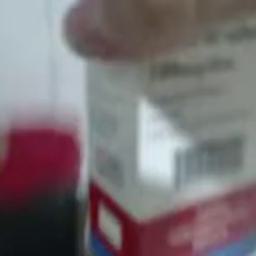}
            \hspace{-1mm}
            \includegraphics[width=0.52in]{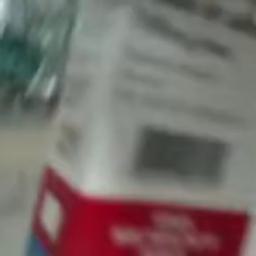}
            \hspace{-1mm}
            \includegraphics[width=0.52in]{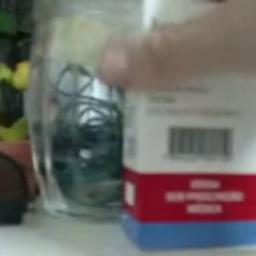}
            \hspace{-1mm}
            \includegraphics[width=0.52in]{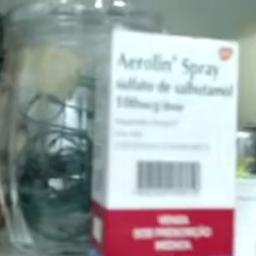}
        \end{minipage}}
    \end{minipage}\\
    \begin{minipage}[b]{1\linewidth}
        \begin{minipage}[b]{0.01\linewidth}
            \rotatebox[origin=c]{90}{Ours}
        \end{minipage} \raisebox{-0.5\height}{
        \begin{minipage}[b]{0.99\linewidth}
            \centering
            \includegraphics[width=0.52in]{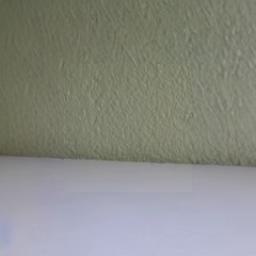}
            \hspace{-1mm}
            \includegraphics[width=0.52in]{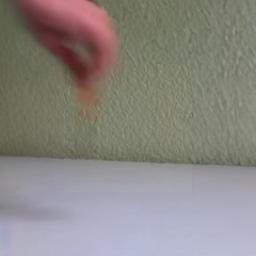}
            \hspace{-1mm}
            \includegraphics[width=0.52in]{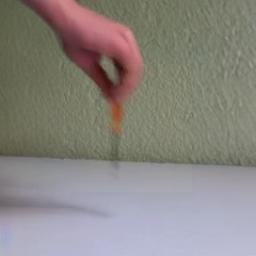}
            \hspace{-1mm}
            \includegraphics[width=0.52in]{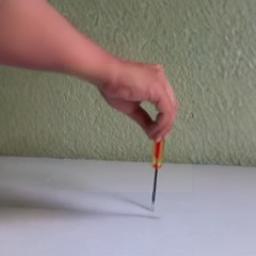}
            \hspace{-1mm}
            \includegraphics[width=0.52in]{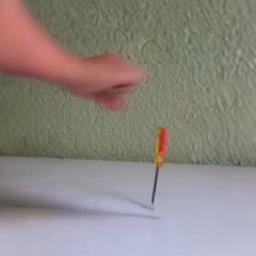}
            \hspace{-1mm}
            \includegraphics[width=0.52in]{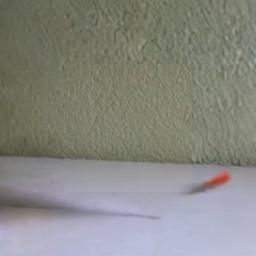}
            \hspace{1mm}
            \includegraphics[width=0.52in]{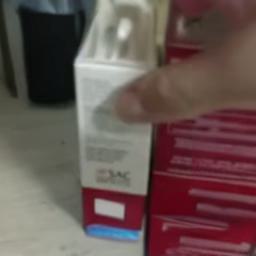}
            \hspace{-1mm}
            \includegraphics[width=0.52in]{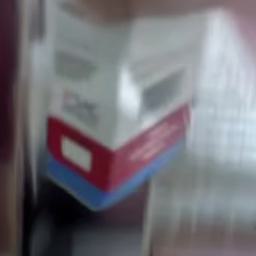}
            \hspace{-1mm}
            \includegraphics[width=0.52in]{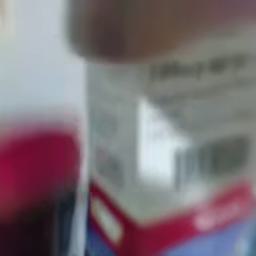}
            \hspace{-1mm}
            \includegraphics[width=0.52in]{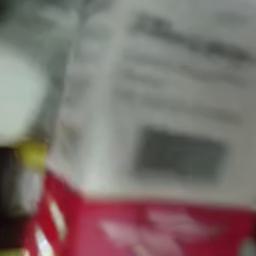}
            \hspace{-1mm}
            \includegraphics[width=0.52in]{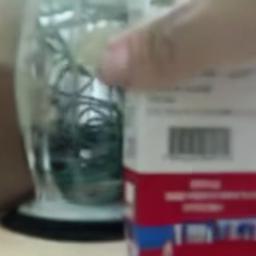}
            \hspace{-1mm}
            \includegraphics[width=0.52in]{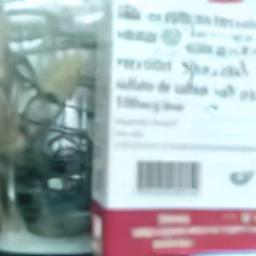}
        \end{minipage}}
    \end{minipage}\\
    \begin{minipage}[b]{1\linewidth}
        \begin{minipage}[b]{0.01\linewidth}
            \rotatebox[origin=c]{90}{GT}
        \end{minipage} \raisebox{-0.5\height}{
        \begin{minipage}[b]{0.99\linewidth}
            \centering
            \includegraphics[width=0.52in]{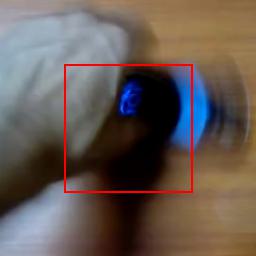}
            \hspace{-1mm}
            \includegraphics[width=0.52in]{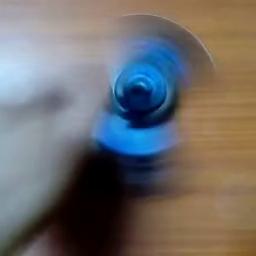}
            \hspace{-1mm}
            \includegraphics[width=0.52in]{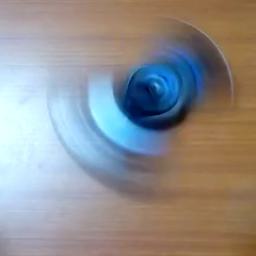}
            \hspace{-1mm}
            \includegraphics[width=0.52in]{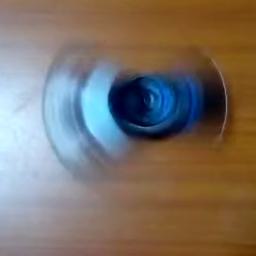}
            \hspace{-1mm}
            \includegraphics[width=0.52in]{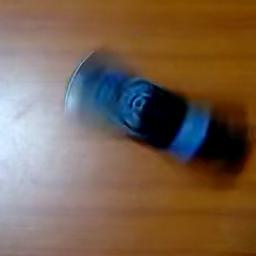}
            \hspace{-1mm}
            \includegraphics[width=0.52in]{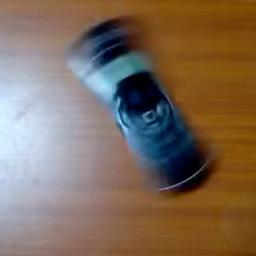}
            \hspace{1mm}
            \includegraphics[width=0.52in]{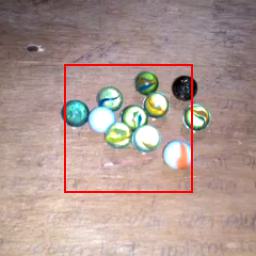}
            \hspace{-1mm}
            \includegraphics[width=0.52in]{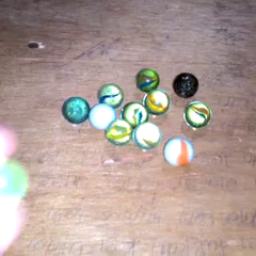}
            \hspace{-1mm}
            \includegraphics[width=0.52in]{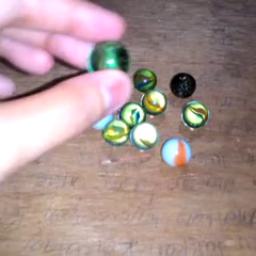}
            \hspace{-1mm}
            \includegraphics[width=0.52in]{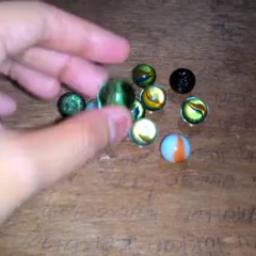}
            \hspace{-1mm}
            \includegraphics[width=0.52in]{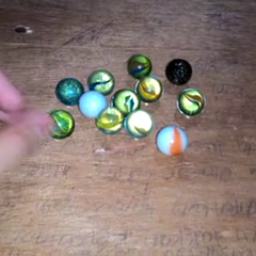}
            \hspace{-1mm}
            \includegraphics[width=0.52in]{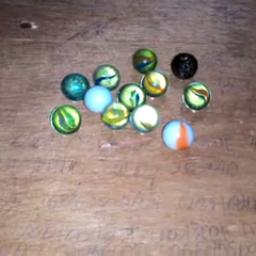}
        \end{minipage}}
    \end{minipage}\\
    \begin{minipage}[b]{1\linewidth}
        \begin{minipage}[b]{0.01\linewidth}
            \rotatebox[origin=c]{90}{Ours}
        \end{minipage} \raisebox{-0.5\height}{
        \begin{minipage}[b]{0.99\linewidth}
            \centering
            \includegraphics[width=0.52in]{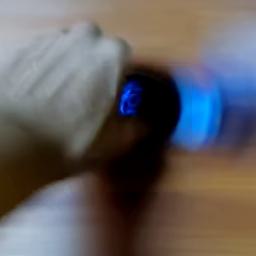}
            \hspace{-1mm}
            \includegraphics[width=0.52in]{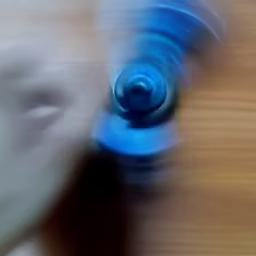}
            \hspace{-1mm}
            \includegraphics[width=0.52in]{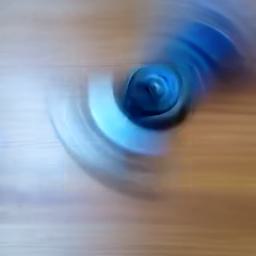}
            \hspace{-1mm}
            \includegraphics[width=0.52in]{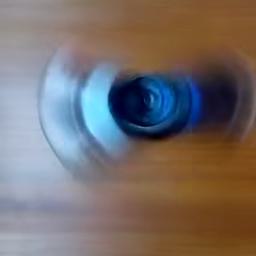}
            \hspace{-1mm}
            \includegraphics[width=0.52in]{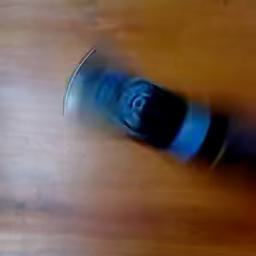}
            \hspace{-1mm}
            \includegraphics[width=0.52in]{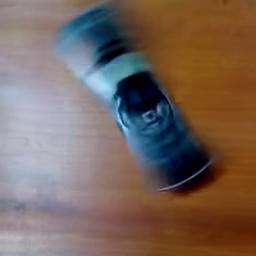}
            \hspace{1mm}
            \includegraphics[width=0.52in]{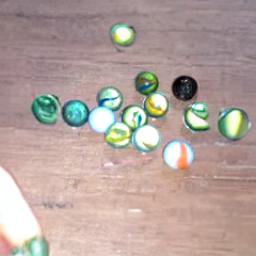}
            \hspace{-1mm}
            \includegraphics[width=0.52in]{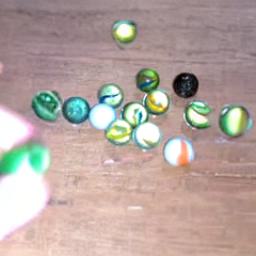}
            \hspace{-1mm}
            \includegraphics[width=0.52in]{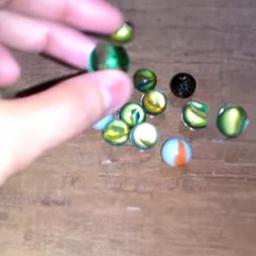}
            \hspace{-1mm}
            \includegraphics[width=0.52in]{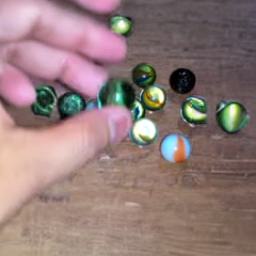}
            \hspace{-1mm}
            \includegraphics[width=0.52in]{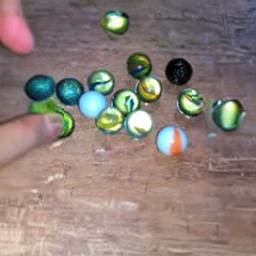}
            \hspace{-1mm}
            \includegraphics[width=0.52in]{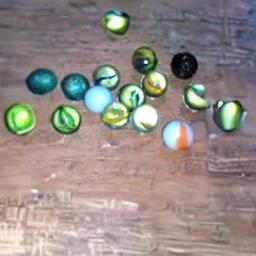}
        \end{minipage}}
    \end{minipage}\\
    \begin{minipage}[b]{1\linewidth}
        \begin{minipage}[b]{0.01\linewidth}
            \rotatebox[origin=c]{90}{GT}
        \end{minipage} \raisebox{-0.5\height}{
        \begin{minipage}[b]{0.99\linewidth}
            \centering
            \includegraphics[width=0.52in]{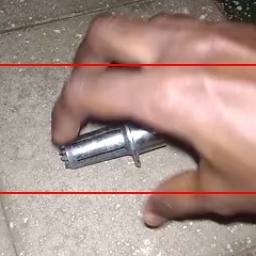}
            \hspace{-1mm}
            \includegraphics[width=0.52in]{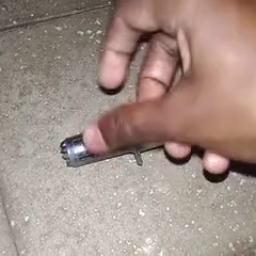}
            \hspace{-1mm}
            \includegraphics[width=0.52in]{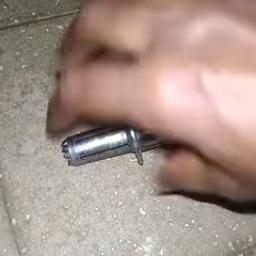}
            \hspace{-1mm}
            \includegraphics[width=0.52in]{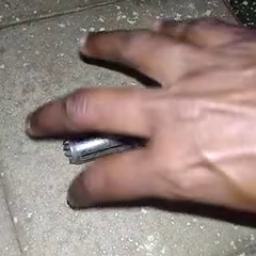}
            \hspace{-1mm}
            \includegraphics[width=0.52in]{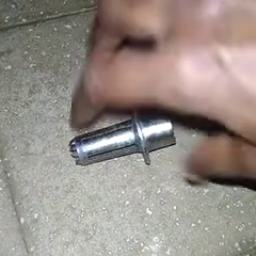}
            \hspace{-1mm}
            \includegraphics[width=0.52in]{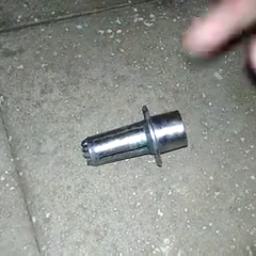}
            \hspace{1mm}
            \includegraphics[width=0.52in]{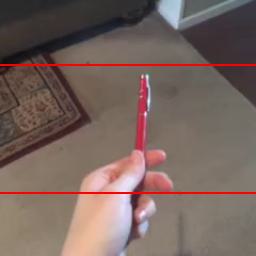}
            \hspace{-1mm}
            \includegraphics[width=0.52in]{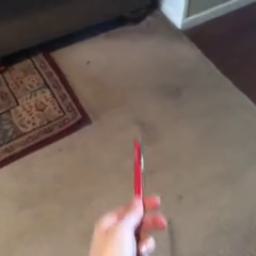}
            \hspace{-1mm}
            \includegraphics[width=0.52in]{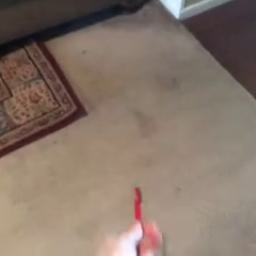}
            \hspace{-1mm}
            \includegraphics[width=0.52in]{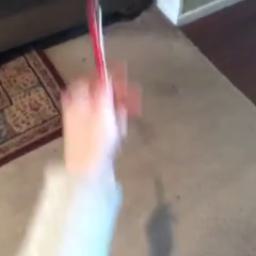}
            \hspace{-1mm}
            \includegraphics[width=0.52in]{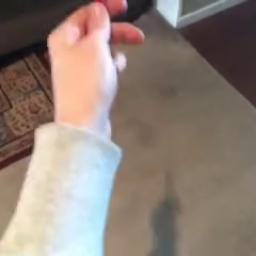}
            \hspace{-1mm}
            \includegraphics[width=0.52in]{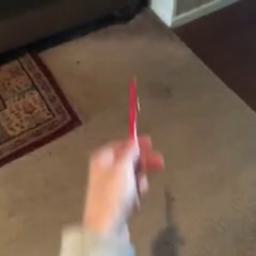}
        \end{minipage}}
    \end{minipage}\\
    \begin{minipage}[b]{1\linewidth}
        \begin{minipage}[b]{0.01\linewidth}
            \rotatebox[origin=c]{90}{Ours}
        \end{minipage} \raisebox{-0.5\height}{
        \begin{minipage}[b]{0.99\linewidth}
            \centering
            \includegraphics[width=0.52in]{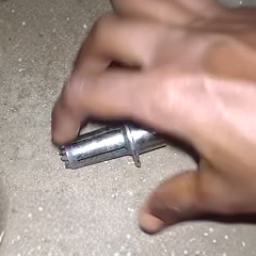}
            \hspace{-1mm}
            \includegraphics[width=0.52in]{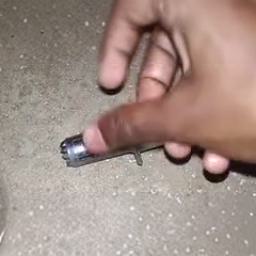}
            \hspace{-1mm}
            \includegraphics[width=0.52in]{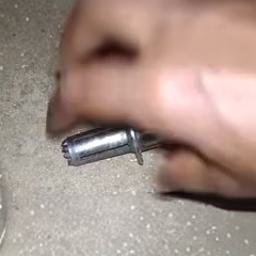}
            \hspace{-1mm}
            \includegraphics[width=0.52in]{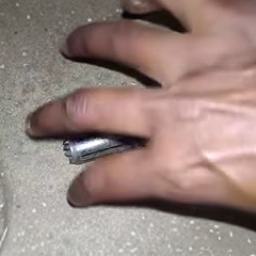}
            \hspace{-1mm}
            \includegraphics[width=0.52in]{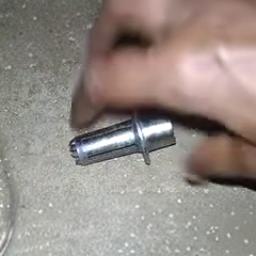}
            \hspace{-1mm}
            \includegraphics[width=0.52in]{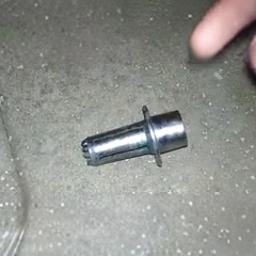}
            \hspace{1mm}
            \includegraphics[width=0.52in]{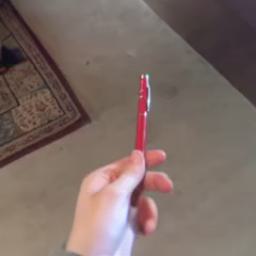}
            \hspace{-1mm}
            \includegraphics[width=0.52in]{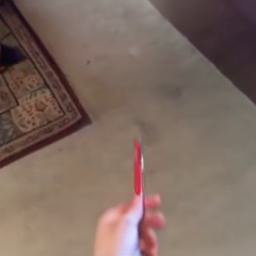}
            \hspace{-1mm}
            \includegraphics[width=0.52in]{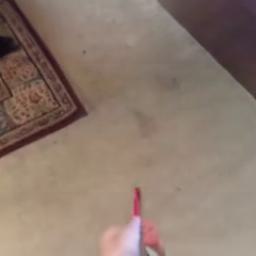}
            \hspace{-1mm}
            \includegraphics[width=0.52in]{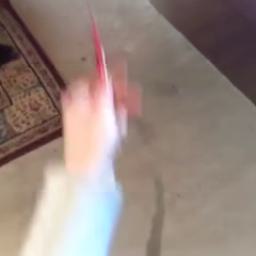}
            \hspace{-1mm}
            \includegraphics[width=0.52in]{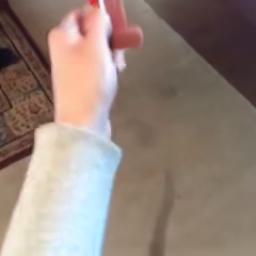}
            \hspace{-1mm}
            \includegraphics[width=0.52in]{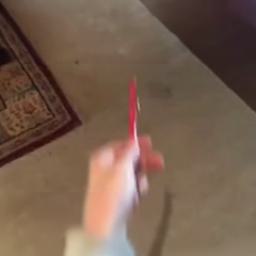}
        \end{minipage}}
    \end{minipage}\\
    \begin{minipage}[b]{1\linewidth}
        \begin{minipage}[b]{0.01\linewidth}
            \rotatebox[origin=c]{90}{GT}
        \end{minipage} \raisebox{-0.5\height}{
        \begin{minipage}[b]{0.99\linewidth}
            \centering
            \includegraphics[width=0.52in]{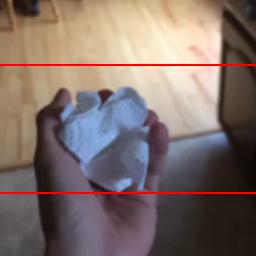}
            \hspace{-1mm}
            \includegraphics[width=0.52in]{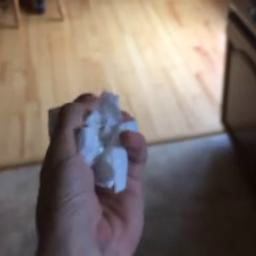}
            \hspace{-1mm}
            \includegraphics[width=0.52in]{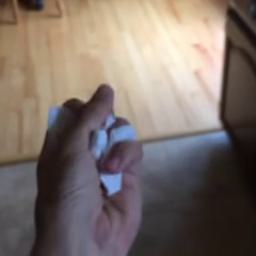}
            \hspace{-1mm}
            \includegraphics[width=0.52in]{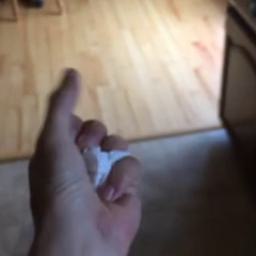}
            \hspace{-1mm}
            \includegraphics[width=0.52in]{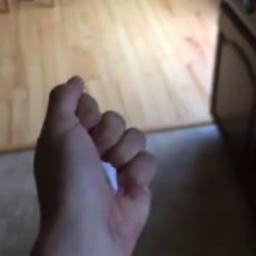}
            \hspace{-1mm}
            \includegraphics[width=0.52in]{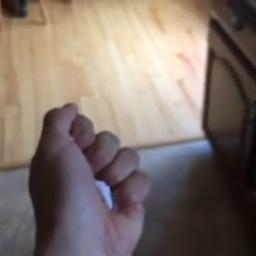}
            \hspace{1mm}
            \includegraphics[width=0.52in]{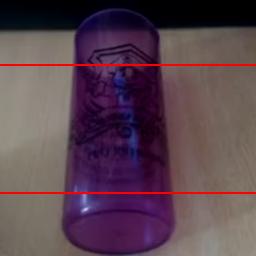}
            \hspace{-1mm}
            \includegraphics[width=0.52in]{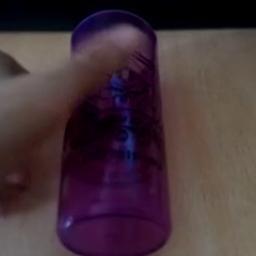}
            \hspace{-1mm}
            \includegraphics[width=0.52in]{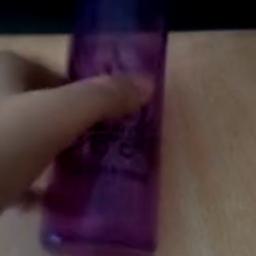}
            \hspace{-1mm}
            \includegraphics[width=0.52in]{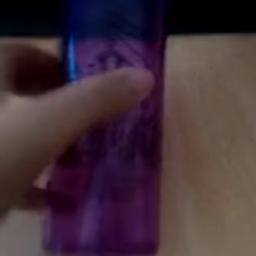}
            \hspace{-1mm}
            \includegraphics[width=0.52in]{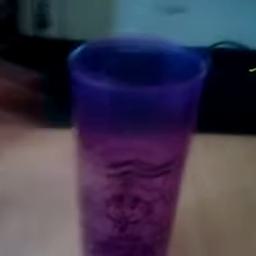}
            \hspace{-1mm}
            \includegraphics[width=0.52in]{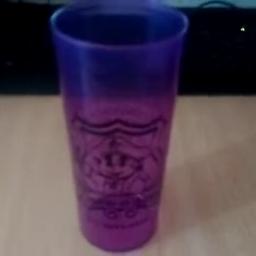}
        \end{minipage}}
    \end{minipage}\\
    \begin{minipage}[b]{1\linewidth}
        \begin{minipage}[b]{0.01\linewidth}
            \rotatebox[origin=c]{90}{Ours}
        \end{minipage} \raisebox{-0.5\height}{
        \begin{minipage}[b]{0.99\linewidth}
            \centering
            \includegraphics[width=0.52in]{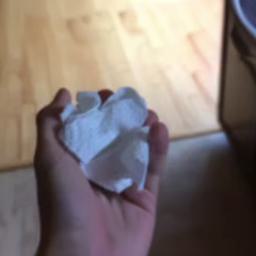}
            \hspace{-1mm}
            \includegraphics[width=0.52in]{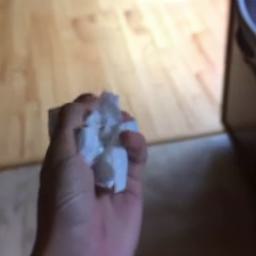}
            \hspace{-1mm}
            \includegraphics[width=0.52in]{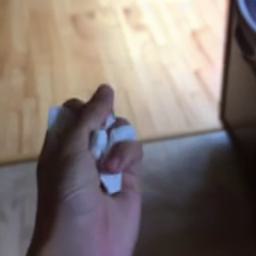}
            \hspace{-1mm}
            \includegraphics[width=0.52in]{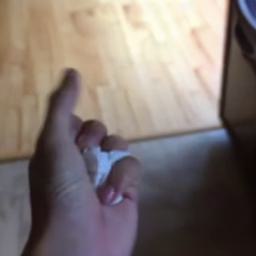}
            \hspace{-1mm}
            \includegraphics[width=0.52in]{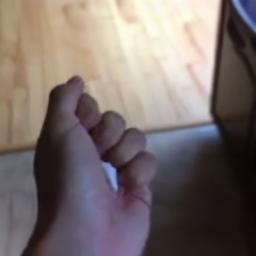}
            \hspace{-1mm}
            \includegraphics[width=0.52in]{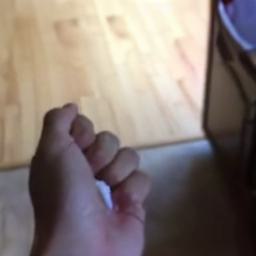}
            \hspace{1mm}
            \includegraphics[width=0.52in]{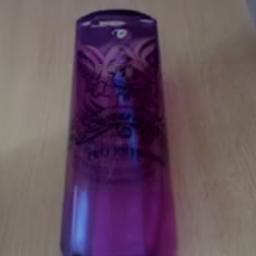}
            \hspace{-1mm}
            \includegraphics[width=0.52in]{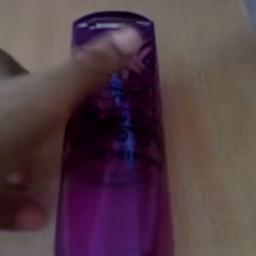}
            \hspace{-1mm}
            \includegraphics[width=0.52in]{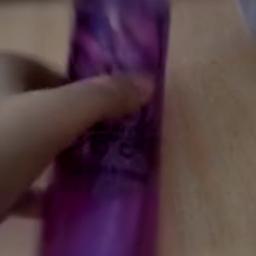}
            \hspace{-1mm}
            \includegraphics[width=0.52in]{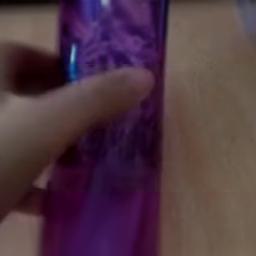}
            \hspace{-1mm}
            \includegraphics[width=0.52in]{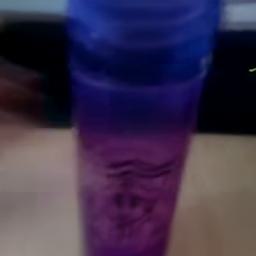}
            \hspace{-1mm}
            \includegraphics[width=0.52in]{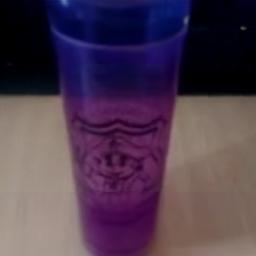}
        \end{minipage}}
    \end{minipage}\\
    \begin{minipage}[b]{1\linewidth}
        \begin{minipage}[b]{0.01\linewidth}
            \rotatebox[origin=c]{90}{GT}
        \end{minipage} \raisebox{-0.5\height}{
        \begin{minipage}[b]{0.99\linewidth}
            \centering
            \includegraphics[width=0.52in]{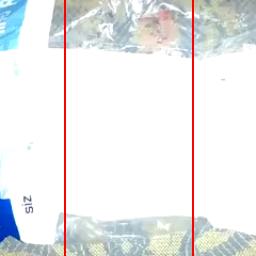}
            \hspace{-1mm}
            \includegraphics[width=0.52in]{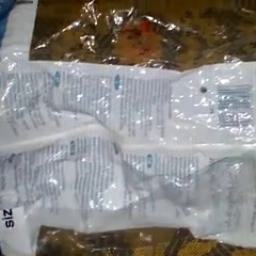}
            \hspace{-1mm}
            \includegraphics[width=0.52in]{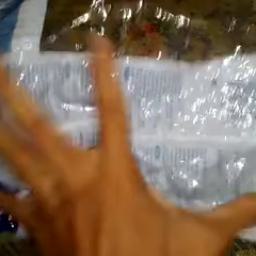}
            \hspace{-1mm}
            \includegraphics[width=0.52in]{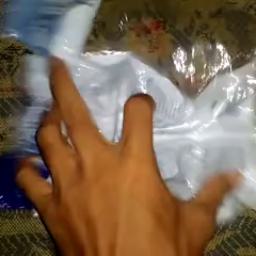}
            \hspace{-1mm}
            \includegraphics[width=0.52in]{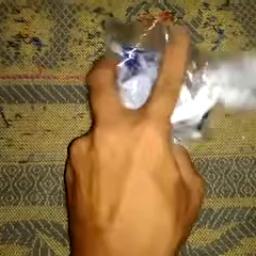}
            \hspace{-1mm}
            \includegraphics[width=0.52in]{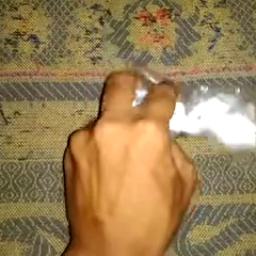}
            \hspace{1mm}
            \includegraphics[width=0.52in]{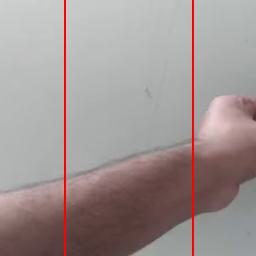}
            \hspace{-1mm}
            \includegraphics[width=0.52in]{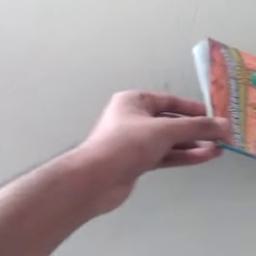}
            \hspace{-1mm}
            \includegraphics[width=0.52in]{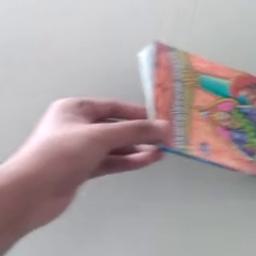}
            \hspace{-1mm}
            \includegraphics[width=0.52in]{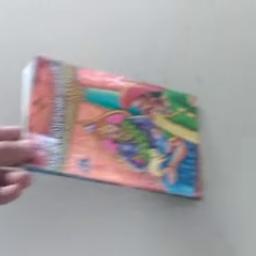}
            \hspace{-1mm}
            \includegraphics[width=0.52in]{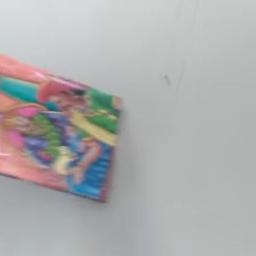}
            \hspace{-1mm}
            \includegraphics[width=0.52in]{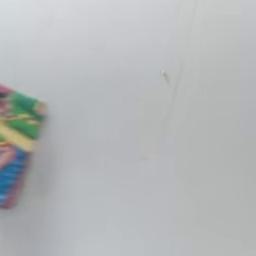}
        \end{minipage}}
    \end{minipage}\\
    \begin{minipage}[b]{1\linewidth}
        \begin{minipage}[b]{0.01\linewidth}
            \rotatebox[origin=c]{90}{Ours}
        \end{minipage} \raisebox{-0.5\height}{
        \begin{minipage}[b]{0.99\linewidth}
            \centering
            \includegraphics[width=0.52in]{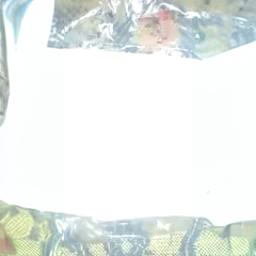}
            \hspace{-1mm}
            \includegraphics[width=0.52in]{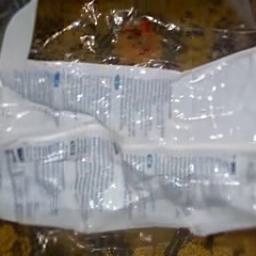}
            \hspace{-1mm}
            \includegraphics[width=0.52in]{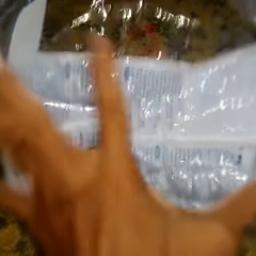}
            \hspace{-1mm}
            \includegraphics[width=0.52in]{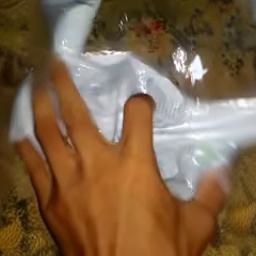}
            \hspace{-1mm}
            \includegraphics[width=0.52in]{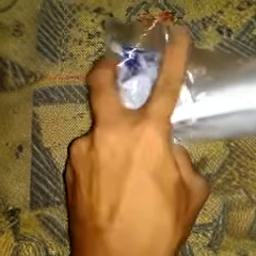}
            \hspace{-1mm}
            \includegraphics[width=0.52in]{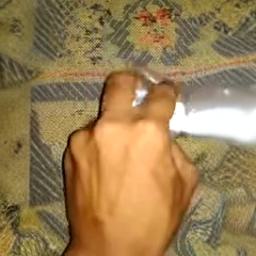}
            \hspace{1mm}
            \includegraphics[width=0.52in]{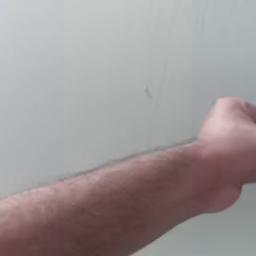}
            \hspace{-1mm}
            \includegraphics[width=0.52in]{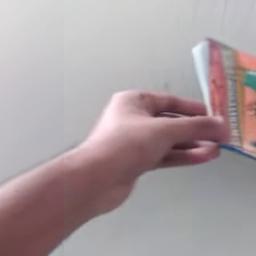}
            \hspace{-1mm}
            \includegraphics[width=0.52in]{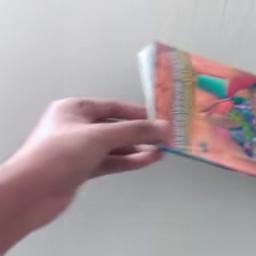}
            \hspace{-1mm}
            \includegraphics[width=0.52in]{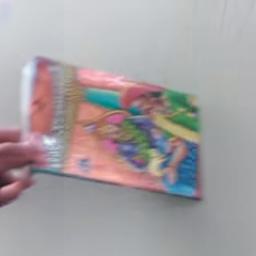}
            \hspace{-1mm}
            \includegraphics[width=0.52in]{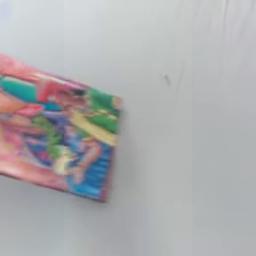}
            \hspace{-1mm}
            \includegraphics[width=0.52in]{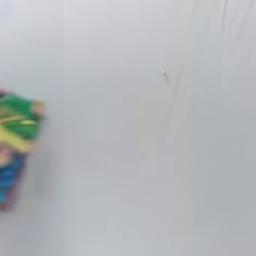}
        \end{minipage}}
    \end{minipage}\\
    \begin{minipage}[b]{1\linewidth}
        \begin{minipage}[b]{0.01\linewidth}
            \rotatebox[origin=c]{90}{GT}
        \end{minipage} \raisebox{-0.5\height}{
        \begin{minipage}[b]{0.99\linewidth}
            \centering
            \includegraphics[width=0.52in]{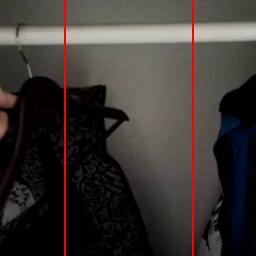}
            \hspace{-1mm}
            \includegraphics[width=0.52in]{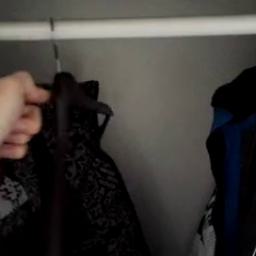}
            \hspace{-1mm}
            \includegraphics[width=0.52in]{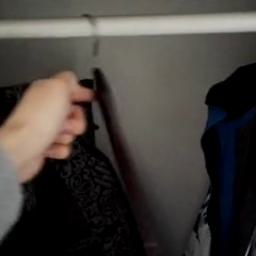}
            \hspace{-1mm}
            \includegraphics[width=0.52in]{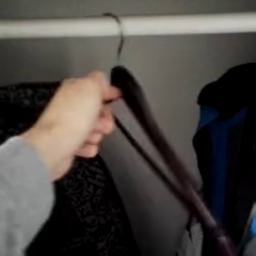}
            \hspace{-1mm}
            \includegraphics[width=0.52in]{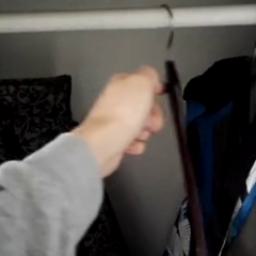}
            \hspace{-1mm}
            \includegraphics[width=0.52in]{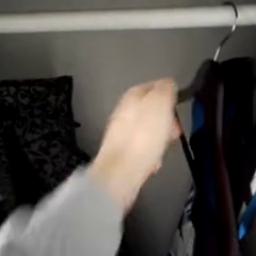}
            \hspace{1mm}
            \includegraphics[width=0.52in]{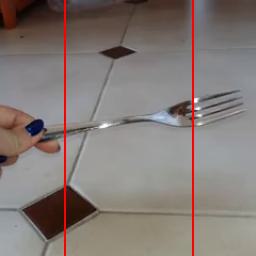}
            \hspace{-1mm}
            \includegraphics[width=0.52in]{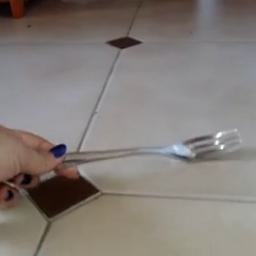}
            \hspace{-1mm}
            \includegraphics[width=0.52in]{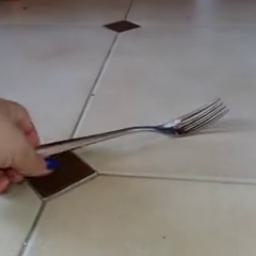}
            \hspace{-1mm}
            \includegraphics[width=0.52in]{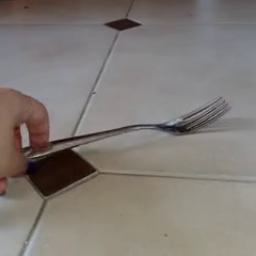}
            \hspace{-1mm}
            \includegraphics[width=0.52in]{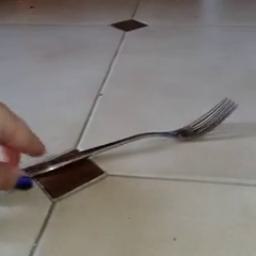}
            \hspace{-1mm}
            \includegraphics[width=0.52in]{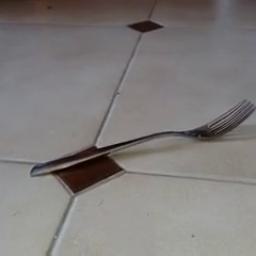}
        \end{minipage}}
    \end{minipage}\\
    \begin{minipage}[b]{1\linewidth}
        \begin{minipage}[b]{0.01\linewidth}
            \rotatebox[origin=c]{90}{Ours}
        \end{minipage} \raisebox{-0.5\height}{
        \begin{minipage}[b]{0.99\linewidth}
            \centering
            \includegraphics[width=0.52in]{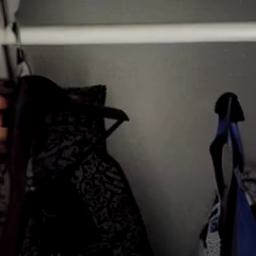}
            \hspace{-1mm}
            \includegraphics[width=0.52in]{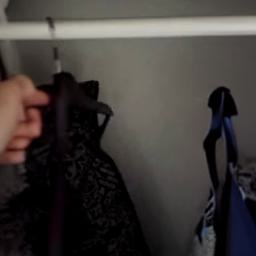}
            \hspace{-1mm}
            \includegraphics[width=0.52in]{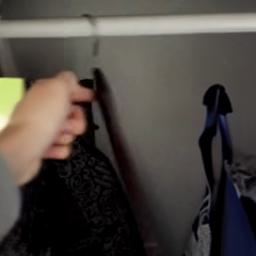}
            \hspace{-1mm}
            \includegraphics[width=0.52in]{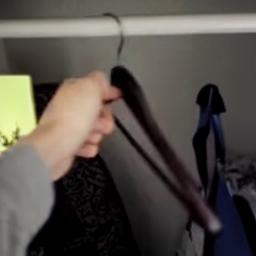}
            \hspace{-1mm}
            \includegraphics[width=0.52in]{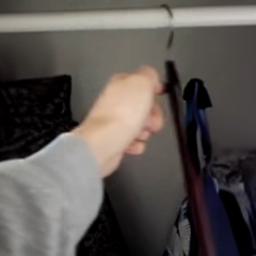}
            \hspace{-1mm}
            \includegraphics[width=0.52in]{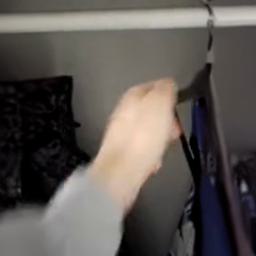}
            \hspace{1mm}
            \includegraphics[width=0.52in]{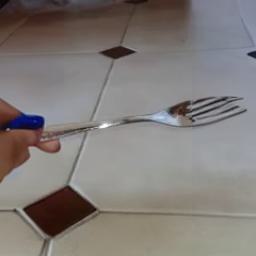}
            \hspace{-1mm}
            \includegraphics[width=0.52in]{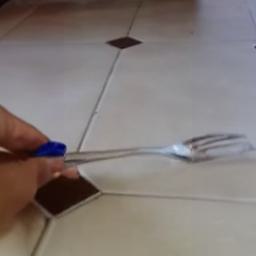}
            \hspace{-1mm}
            \includegraphics[width=0.52in]{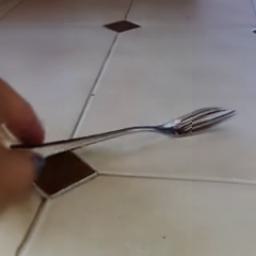}
            \hspace{-1mm}
            \includegraphics[width=0.52in]{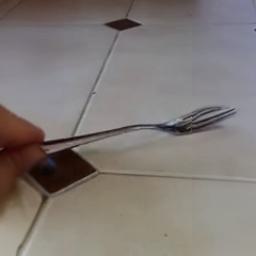}
            \hspace{-1mm}
            \includegraphics[width=0.52in]{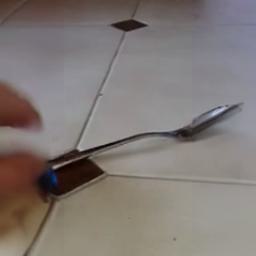}
            \hspace{-1mm}
            \includegraphics[width=0.52in]{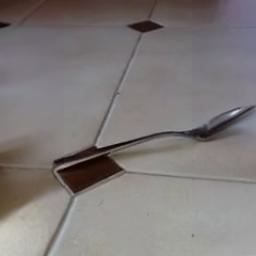}
        \end{minipage}}
    \end{minipage}
    \caption{Three types of video outpainting on the SSv2 dataset.
    The term GT refers to ground truth, and for each set of GT, the area to be filled is marked with red curves on the first image~(the area outside the red lines is what we want to fill in).}\label{fig:ssv2res}
\end{figure*}

In order to compare with the MAGVIT~\cite{yu2022magvit}, we obtain the evaluation results directly from their paper. They evaluated three types of video outpainting FVD~\cite{unterthiner2018towards} scores on the Something-Something V.2~(SSv2)~\cite{goyal2017something,mahdisoltani2018effectiveness} dataset. The three types of outpainting are Central Outpainting (OPC), Vertical Outpainting (OPV), and Horizontal Outpainting (OPH). The mask ratio for each type is 0.75 for OPC, 0.5 for OPV, and 0.5 for OPH. We strictly follow their setup, using 169K videos for training and 24K videos for evaluation on the SSv2 dataset. We train the dataset using 24 A100 GPUs, with a batch size of 240 and fine-tuned for 126k steps. The average video length of SSv2 is around 30 frames, and we use the dense prediction, following the settings of short video outpainting in the main paper we reported. We use the same FVD~\cite{unterthiner2018towards} evaluation metric as them, with 16 frames for each video. Each evaluated video is sampled with 2 temporal windows and a central crop with a frame size of 128. The comparison results are shown in Table~\ref{tab:quantitativemagvit}. We also present the qualitative results of the three types of video outpainting in Fig.~\ref{fig:ssv2res}.

\section{Network Architecture and Implementation Details}

\subsection{Network Architecture}
\label{sec:netarch}
Our approach consists of two trainable networks: a 3D denoising Unet and a lightweight video encoder.
Our 3D denoising UNet uses the pre-trained parameters from the text-to-image model in LDMs.
In order to adapt it for our task with a 3D structure, we employ temporal convolution, self-attention, and cross-attention operations to ensure the interaction between different frames.
Our 3D denoising Unet takes latents from the VAE encoder~\cite{rombach2022high} as input, with dimensions of $(batch\_size, num\_frames\_of\_video, in\_channels, height, \\ weight)$.
Our 3D denoising Unet predicts the noise with shape $(batch\_size, num\_frames\_of\_video, out\_channels, height, weight)$.
In our implementation, $in\_channels$ is 9, where 8 dimensions represent the latent of the original video frames and masked frames (with 4 dimensions each), and 1 dimension represents the mask.
$out\_channels$ is 4, the same as the latent of the original video frames.
After compression by VAE, the dimensions of our height and weight become 32.
Our 3D denoising UNet heavily references the network structure in Make-A-Video~\cite{singer2022make}.
We follow the Make-A-Video~\cite{singer2022make} by utilizing Pseudo-3D convolutional and attention layers to leverage pre-trained text-to-image models within the latent diffusion models(LDMs)~\cite{rombach2022high}.
Each spatial 2D conv layer is followed by a temporal 1D conv layer.
We not only add the timestep embeddings of the noise to each layer but also add the fps rate embeddings.
This allows us to use one model to generate video clips with different frame intervals.
Our 3D denoising Unet has four downsampling and four upsampling layers, with each layer outputting the following number of channels: [320, 640, 1280, 1280].
Our 3D Unet has a total of 1299.28M parameters.
For more details, we recommend referring to the network architecture in Make-A-Video~\cite{singer2022make}.

\begin{figure}
    \includegraphics[width=\linewidth]{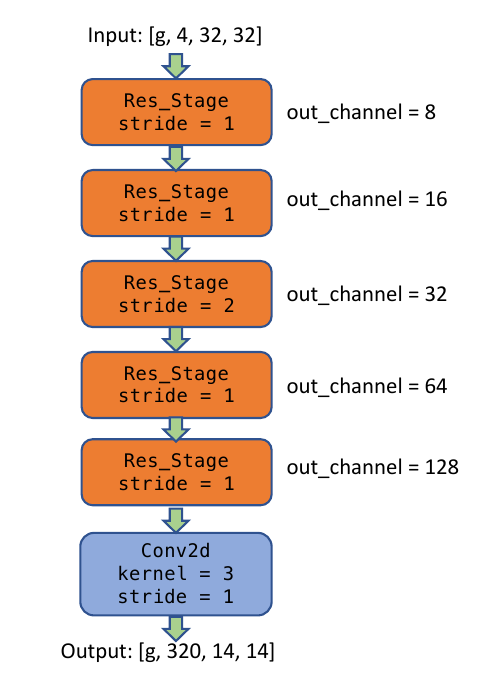}
    \caption{Our lightweight video encoder. $g$ denotes the total number of global video frames inputted. In our implementation $g = 16$.
    We referred to the image encoder in designbooster~\cite{sun2023design}.}\label{fig:net2}
\end{figure}

We have presented our lightweight video encoder in Fig.~\ref{fig:net2}.
Our lightweight video encoder accepts the global video latents obtained from VAE and increases its dimensionality from 4 to 320 for cross-attention.

\input{table/compare_magvit}

\subsection{Implementation Details}
\label{sec:implementation}

\textbf{Sampling Details.} We use the PNDMScheduler from pseudo numerical methods for diffusion models (PNDMs)~\cite{liu2022pseudo}. We use 50 inference steps and a scaled linear $\beta$ schedule that starts at 0.00085 and ends at 0.012.

Our 3D denoising UNet is capable of generating $F=16$ frames in a single inference, and we use $g=16$ global frames.
we randomly extract F frames from video clips, with equal intervals between each frame.
The frame intervals are uniformly sampled from fps $[1, 30]$. 
We employ the Adam~\cite{kingma2014adam} optimizer with a learning rate of 1e-4, and the warm-up learning rate step is 1k.
We trained the model for 4 epochs on the WebVid dataset~\cite{Bain21} and then fine-tuned it for 3 epochs on our 5M e-commerce dataset.
All training was done on 24 A100 GPUs, and the entire training process took approximately 2.5 weeks.
We use the dense predict form for short video outpainting and the three-level coarse-to-fine structure with time intervals of [30, 15, 1] for long video outpainting.
We found that the inference methods with frame intervals of [15, 5, 1] were nearly equally effective.
However, considering the length of our long videos, we opted for the inference method with frame intervals of [30, 15, 1].
We set $s_{1}=2$ and $s_{2}=4$ because experiments show that this leads to good outpainting results.

The resolution of our input video is 256 x 256 x 3. During the test phase, we can infer test samples with a batch size of 2 on a 16GB graphics card (the test environment we use is Tesla v100 16Gb). Our training phase used 24 80GB A100 GPUs, with a total batch size of 240.

\section{Limitations and bad cases}

\begin{figure}
    \includegraphics[width=0.6in]{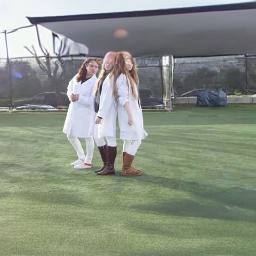}
    \includegraphics[width=0.6in]{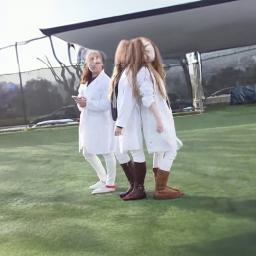}
    \includegraphics[width=0.6in]{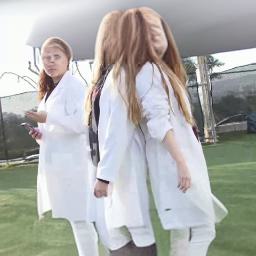}
    \includegraphics[width=0.6in]{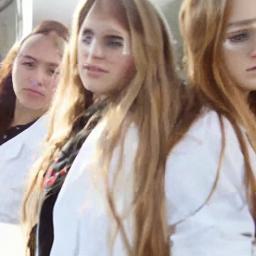}
    \includegraphics[width=0.6in]{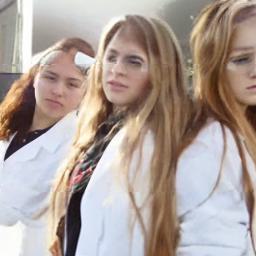}\\
    \includegraphics[width=0.6in]{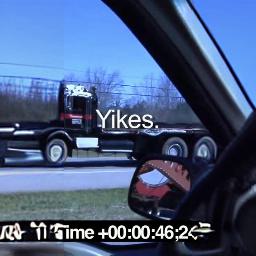}
    \includegraphics[width=0.6in]{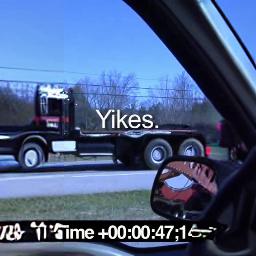}
    \includegraphics[width=0.6in]{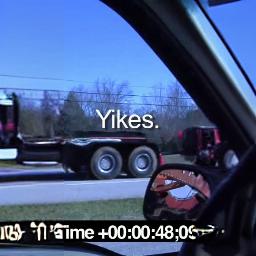}
    \includegraphics[width=0.6in]{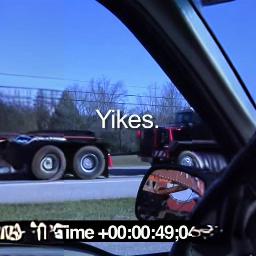}
    \includegraphics[width=0.6in]{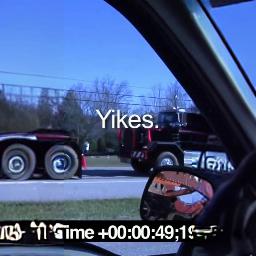} \\
    \includegraphics[width=0.6in]{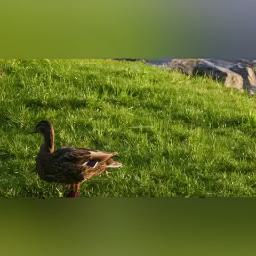}
    \includegraphics[width=0.6in]{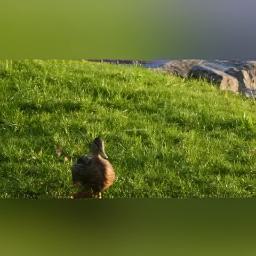}
    \includegraphics[width=0.6in]{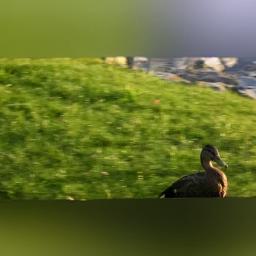}
    \includegraphics[width=0.6in]{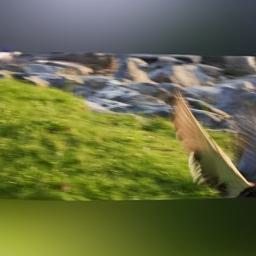}
    \includegraphics[width=0.6in]{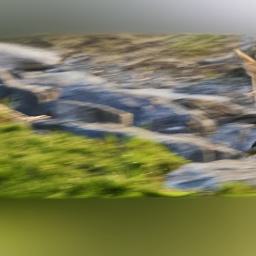}\\
    \includegraphics[width=0.6in]{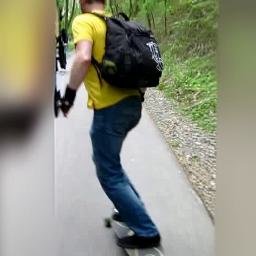}
    \includegraphics[width=0.6in]{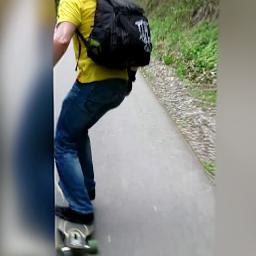}
    \includegraphics[width=0.6in]{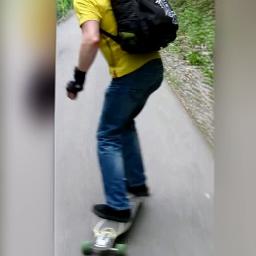}
    \includegraphics[width=0.6in]{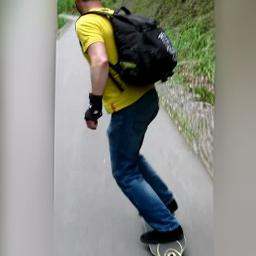}
    \includegraphics[width=0.6in]{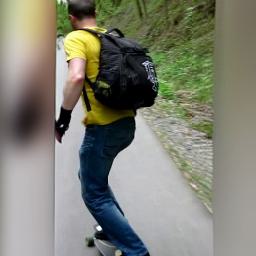}
    \caption{Bad case in our method.}\label{ourbad1}
\end{figure}

We show the bad cases generated by our model in Fig.~\ref{ourbad1}.
Our method utilizes a fixed image VAE~\cite{rombach2022high} encoder to transform the pixel-space video into the latent space.
VAE often shows rough performance in human faces and some fine structures.
Moreover, our method is limited by the training data and the difficulty of the problem, resulting in poor results in text generation within videos.

Our diffusion model is sensitive to the initial Gaussian noise during sampling, and some videos may experience edge blurring.
We have performed a simple preprocessing step on the extended region of the video to be predicted using the OpenCV inpaint function and added 1000 steps of Gaussian noise instead of directly sampling from the Gaussian noise, which partially solves the problem of prediction robustness.

\end{document}

%% file: table/compare_davis_youtube.tex
\begin{table*}[t]
\caption{Quantitative evaluation of video outpainting on the DAVIS and YouTube-VOS datasets.}\label{tab:quantitative1}
    \begin{center}
        \begin{tabular}{ccccccccccc}
            \hline
            \multirow{2}{*}{Method} & \multicolumn{5}{c}{Davis dataset \cite{perazzi2016benchmark}} &  \multicolumn{5}{c}{YouTube-VOS dataset \cite{xu2018large}}\\
            & MSE $\downarrow$ & PSNR $\uparrow$ & SSIM $\uparrow$ & LPIPS $\downarrow$ & FVD $\downarrow$ &MSE $\downarrow$ & PSNR $\uparrow$ & SSIM $\uparrow$ & LPIPS $\downarrow$ & FVD $\downarrow$ \\
            \hline
            Dehan \cite{dehan2022complete} & 0.0260           & 17.96          & 0.6272          & 0.2331          & 363.1 &0.02312         &18.25          &0.7195          &0.2278 & 149.7 \\
            SDM   \cite{rombach2022high}   & 0.0153          & 20.02          & 0.7078 & 0.2165          & 334.6 & 0.01687         &19.91          &0.7277          &0.2001 & 94.81 \\
            Ours                           & \textbf{0.0149} & \textbf{20.26} & \textbf{0.7082}          & \textbf{0.2026} & \textbf{300.0} &\textbf{0.01636}&\textbf{20.20} &\textbf{0.7312} &\textbf{0.1854} &\textbf{66.62} \\
            \hline
        \end{tabular}
    \end{center}

\end{table*}

%% file: table/ablation.tex
\begin{table}[t]
\caption{Ablation study on our e-commerce dataset. `w/o' means without.}\label{tab:ablation} 
    \begin{center}
        \setlength{\tabcolsep}{1mm}
        \begin{tabular}{ccccccccccc}
            \toprule
            Method                        &  MSE $\downarrow$             & PSNR   $\uparrow$        & SSIM $\uparrow$           & LPIPS   $\downarrow$         & FVD $\downarrow$ \\
            \hline
            \small{SDM}                   & 0.01134           & 17.92          & 0.6783          & 0.2139           & 110.4 \\
            \small{MSDM w/o prompt}            & 0.00914           & 19.22          & 0.6912          & 0.2012           & 70.8 \\
            \small{Ours} & \textbf{0.00791}  & \textbf{20.01} & \textbf{0.7112} & \textbf{0.1931}  & \textbf{68.3} \\
            \bottomrule
        \end{tabular}
    \end{center}

\end{table}

%% file: table/compare_magvit.tex
\begin{table}
  \caption{Evaluate the performance of video outpainting using FVD on something-something-v2. We obtain the results directly from MAGVIT.}~\label{tab:quantitativemagvit}
  \begin{tabular}{cccccc}
    \toprule
    Method & AVG & OPC & OPV & OPH \\
    \midrule
    MAGVIT-L-MT~\cite{yu2022magvit} & 18.3 & 21.1 & 16.8 & 17.0 \\
    Ours & \textbf{16.0}  & \textbf{19.2}   &  \textbf{14.5}   & \textbf{14.3}  \\
    \bottomrule
  \end{tabular}
\end{table}